\documentclass{article} 
\usepackage[T1]{fontenc}
\usepackage{iclr2025_conference,times}

\usepackage{amsmath,amsfonts,bm}

\def\eqref#1{equation~\ref{#1}}

\def\1{\bm{1}}

\DeclareMathAlphabet{\mathsfit}{\encodingdefault}{\sfdefault}{m}{sl}
\SetMathAlphabet{\mathsfit}{bold}{\encodingdefault}{\sfdefault}{bx}{n}

\usepackage[breaklinks=true]{hyperref}
\usepackage{breakurl}
\usepackage{graphicx}
\usepackage{wrapfig}
\usepackage{placeins}
\usepackage{booktabs}
\usepackage{multirow}
\usepackage{enumitem}
\usepackage{float}
\usepackage{tabularx}
\PassOptionsToPackage{table,usenames,dvipsnames}{xcolor}
\usepackage[most]{tcolorbox}
\usepackage{lipsum}
\usepackage{anyfontsize}
\usepackage{xspace}
\usepackage{microtype}
\usepackage[marginpar]{todo}
\usepackage{tikz}
\usepackage{pgfplots}
\usepackage{makecell}
\pgfplotsset{compat=1.18}

\definecolor{glaucous}{rgb}{0.38, 0.51, 0.71}
\definecolor{pastelmagenta}{rgb}{0.96, 0.6, 0.76}
\definecolor{orchid}{rgb}{0.85, 0.44, 0.84}
\definecolor{postTrainingColor}{RGB}{155, 130, 190}

\hypersetup{
  colorlinks=true,
  citecolor=glaucous,
  linkcolor=pastelmagenta,
  urlcolor=orchid}

\usepackage{arydshln}
\makeatletter
\def\adl@drawiv#1#2#3{%
        \hskip.5\tabcolsep
        \xleaders#3{#2.5\@tempdimb #1{1}#2.5\@tempdimb}%
                #2\z@ plus1fil minus1fil\relax
        \hskip.5\tabcolsep}
\newcommand{\cdashlinelr}[1]{%
  \noalign{\vskip 1.3pt
           \global\let\@dashdrawstore\adl@draw
           \global\let\adl@draw\adl@drawiv}
  \cdashline{#1}[.4pt/2pt]
  \noalign{\global\let\adl@draw\@dashdrawstore
           \vskip 3pt}}
\makeatother

\usepackage{marvosym}

\title{\centering EuroLLM-22B: Technical Report}

\author{
\vspace{0.3cm}
\bf
Miguel Moura Ramos*$^{1,2}$ \hspace{0.1cm}
Duarte M. Alves*$^{1,2}$ \hspace{0.1cm}
Hippolyte Gisserot-Boukhlef*$^{3,12}$ \\\bf
João Alves\textsuperscript{\Neptune}$^{4}$ \hspace{0.1cm}
Pedro Henrique Martins\textsuperscript{\Neptune}$^{10}$ \hspace{0.1cm}
Patrick Fernandes$^{1,2,5}$ \hspace{0.1cm}
José Pombal\textsuperscript{\Neptune}$^{1,2,10}$ \\\bf
Nuno M. Guerreiro\textsuperscript{\Neptune}$^{10}$ \hspace{0.1cm}
Ricardo Rei\textsuperscript{\Neptune}$^{10}$ \hspace{0.1cm}
Nicolas Boizard$^{3,7}$ \hspace{0.1cm}
Amin Farajian\textsuperscript{\Neptune}$^{13}$ \\\bf
Mateusz Klimaszewski$^{6}$ \hspace{0.1cm}
José G. C. de Souza\textsuperscript{\Neptune}$^{9}$ \hspace{0.1cm}
Barry Haddow$^{6,8}$  \hspace{0.1cm}
François Yvon$^{11}$ \\\bf
Pierre Colombo$^{3}$ \hspace{0.1cm}
Alexandra Birch$\diamond^{6,8}$ \hspace{0.1cm}
André F. T. Martins\textsuperscript{\Neptune}$\diamond^{1,2,13}$
\\
\vspace{0.3cm}
\normalfont
$^{1}$ Instituto Superior Técnico \& Universidade de Lisboa (Lisbon ELLIS Unit) \\ 
$^{2}$Instituto de Telecomunicações \hspace{0.1cm} $^{3}$MICS, CentraleSupélec, Université Paris-Saclay \hspace{0.1cm} $^{4}$Acolad \\ 
$^{5}$Carnegie Mellon University \hspace{0.1cm} $^{6}$University of Edinburgh \hspace{0.1cm} $^{7}$Diabolocom \hspace{0.1cm} $^{8}$Aveni \hspace{0.1cm}  $^{9}$OutSystems \\ 
$^{10}$Sword Health \hspace{0.1cm} \hspace{0.1cm} $^{11}$Sorbonne Université, CNRS, ISIR \hspace{0.1cm} $^{12}$Artefact Research Center \hspace{0.1cm} $^{13}$TransPerfect
}

\newcommand{\eurollm}{EuroLLM\xspace{}}
\newcommand{\euroblocks}{EuroBlocks\xspace{}}
\newcommand{\eurofilter}{EuroFilter\xspace{}}

\newcommand\blfootnote[1]{%
  \begingroup
  \renewcommand\thefootnote{}\footnote{#1}%
  \addtocounter{footnote}{-1}%
  \endgroup
}

\iclrfinalcopy

\begin{document}

\maketitle

\begin{abstract}
\blfootnote{* Core contributors, $\diamond$ equal contributors, \Neptune\space work done while working at Unbabel.}%
This report presents \emph{EuroLLM-22B}, a large language model trained from scratch to support the needs of European citizens by covering all 24 official European Union languages and 11 additional languages. EuroLLM addresses the issue of European languages being underrepresented and underserved in existing open large language models. We provide a comprehensive overview of EuroLLM-22B's development, including tokenizer design, architectural specifications, data filtering, and training procedures.\footnote{All resources are available on the HuggingFace portal as part of the \href{https://huggingface.co/collections/utter-project/eurollm-66b2bd5402f755e41c5d9c6d}{\eurollm{} Collection}.}
Across a broad set of multilingual benchmarks, EuroLLM-22B demonstrates strong performance in reasoning, instruction following, and translation, achieving results competitive with models of comparable size.
To support future research, we release our base and instruction-tuned models, our multilingual web pretraining data and updated \euroblocks{} instruction datasets, as well as our pre-training and evaluation codebases.
\end{abstract}

\section{Introduction}

Large language models~(LLMs) continue to drive progress in natural language processing, pushing substantial advances in reasoning, multilinguality, and instruction following \citep{wei-etal-2022-chain,ouyang-etal-2022-training,deepseekai2025deepseekv3technicalreport}.
Despite these developments, most leading models are either closed \citep{ClaudeThree,openai2024gpt4technicalreport,comanici2025gemini25pushingfrontier} or only partially open---commonly releasing model weights but providing limited transparency about training data or procedures \citep{dubey2024llama,yang2025qwen3,team2025gemma,team2025kimi}. 
While fully open alternatives do exist \citep{olmo2025olmo}, they often prioritise English or a small set of high-resource languages.
As a result, in the current open model ecosystem, many European languages remain underserved~\citep{rehm2023language}  and relatively few LLMs have been ``made in Europe'' \citep{bigscience-etal-2022-bloom,jiang2024mixtral,gonzalezagirre2025salamandratechnicalreport,hernándezcano2025apertus}. 

We launched the \href{https://eurollm.io}{\eurollm{}} project to address this gap by developing open models that natively support all 24 official European Union (EU) languages, fostering the development of AI technologies in the EU.
Our earlier releases, \eurollm{} 1.7B~\citep{martins2024eurollm} and \eurollm{} 9B~\citep{martins-etal-2025-eurollm9b}, demonstrated strong multilingual capabilities and competitive translation performance when compared to existing open alternatives, marking important progress toward this objective. 
Overall, \eurollm{} supports the 24 official EU languages 
(Bulgarian, Croatian, Czech, Danish, Dutch, English, Estonian, Finnish, French, German, Greek, Hungarian, Irish, Italian, Latvian, Lithuanian, Maltese, Polish, Portuguese, Romanian, Slovak, Slovenian, Spanish, and Swedish) and 11 additional languages (Arabic, Catalan, Chinese, Galician, Hindi, Japanese, Korean, Norwegian, Russian, Turkish, and Ukrainian). 

Building on this trajectory, we introduce \eurollm{} 22B, our largest and most capable model to date.
For this release, we improve the quality of the pre-training corpus through large-scale multilingual data filtering, adopting a multi-phase training strategy that progressively exposes the model to higher-quality data.
We further extend the context window to 32K tokens, enabling more effective modeling of long-form inputs.
In addition, we substantially expand and strengthen the post-training data by introducing a new version of \euroblocks{}, a multilingual instruction dataset constructed from diverse public sources and enhanced with higher-quality synthetic responses \citep{NemotronPostTrainingDatasetV1,NemotronPostTrainingDatasetV2,teknium2024hermes3technicalreport,lambert2025tulu}.
Together, these improvements yield significant gains in multilingual reasoning and instruction-following performance.
Across a wide range of multilingual benchmarks, \eurollm{} 22B achieves competitive results relative to leading open models of similar scale, positioning it as a highly capable model of its size.

Along with this technical report, we release:
\begin{itemize}[leftmargin=11pt, itemsep=3pt, parsep=0pt, topsep=0pt]
    \item \textbf{Instruct models}: the \href{https://huggingface.co/utter-project/EuroLLM-22B-Instruct-2512}{EuroLLM-22B} model, together with an improved \href{https://huggingface.co/utter-project/EuroLLM-9B-Instruct-2512}{EuroLLM-9B} obtained adopting the same post-training recipe as \eurollm{}-22B;
    \item \textbf{Base models}: the \href{https://huggingface.co/utter-project/EuroLLM-22B-2512}{EuroLLM-22B-Base} model, together with an improved \href{https://huggingface.co/utter-project/EuroLLM-9B-2512}{EuroLLM-9B-Base} version adopting the same long context extension (32K) as \eurollm{}-22B-Base;
    \item \textbf{Data}: the \href{https://huggingface.co/datasets/utter-project/EuroLLM-Multilingual-Data-2512}{EuroWeb} dataset, our multilingual web dataset used for pre-training \eurollm{}~22B, together with a new version of \href{https://huggingface.co/datasets/utter-project/EuroBlocks-SFT-2512}{\euroblocks{}}, our multilingual instruction dataset which we used in the post-training our models;
    \item \textbf{Open-source code}: our \href{https://github.com/deep-spin/Megatron-LM-pretrain}{fork} of Megatron-LM~\citep{megatron-lm} for pretraining, and \href{https://github.com/deep-spin/eurollm-eval}{code} to reproduce all model evaluations.
\end{itemize}

\section{Pre-training}
\label{sec:pretraining}

We first describe the modeling and architectural design of EuroLLM-22B (\S\ref{subsec:modeling}), then outline the multi-phase training procedure (\S\ref{subsec:training_phases}), and finally detail the composition and curation of the pre-training dataset (\S\ref{subsec:dataset}). We pretrain our models using NVIDIA's Megatron-LM~\citep{megatron-lm}, which we extend to support our scheduler.\footnote{\url{https://github.com/deep-spin/Megatron-LM-pretrain}.}

\subsection{Modeling}
\label{subsec:modeling}

\begin{table}[t]
\small
\centering
\begin{tabular}{lccc}
\toprule
& \textbf{1.7B} & \textbf{9B} & \textbf{22B} \\ \midrule
Sequence Length & 4,096 & 4,096  & 32,768 \\
Number of Layers & 24 & 42 & 54 \\
Embedding Size & 2,048 & 4,096 & 6,144 \\
FFN Hidden Size & 5,632 & 12,288 & 16,384 \\
Number of Heads & 16 & 32 & 48 \\
Number of KV Heads (GQA) & 8 & 8 & 8 \\
Activation Function & SwiGLU & SwiGLU & SwiGLU \\
Position Encodings & RoPE ($\Theta=1\times 10^{4})$ & RoPE ($\Theta=1\times 10^{4})$ & RoPE ($\Theta=1\times 10^{6})$ \\
Layer Norm & RMSNorm & RMSNorm & RMSNorm \\
Tied Embeddings & No & No & No \\
Max Learning Rate & $3\times 10^{-4}$ & $3\times 10^{-4}$ & $3\times 10^{-4}$ \\
Min Learning Rate & $3\times 10^{-5}$ & $3\times 10^{-5}$ & $3\times 10^{-5}$ \\
\midrule
Embedding Parameters & 0.262B & 0.524B & 0.768B \\
LM Head Parameters & 0.262B & 0.524B & 0.768B \\
Non-embedding Parameters & 1.133B & 8.105B & 21.067B \\
Total Parameters & 1.657B & 9.153B & 22.639B \\
\bottomrule
\end{tabular} 
\caption{EuroLLM hyperparameters for the 1.7B, 9B, and 22B models, for comparison purposes.}
\label{tab:hyperparameters}
\end{table}

\eurollm{} 22B follows most of the design decisions made during the development of the 1.7B~\citep{martins2024eurollm} and the 9B~\citep{martins-etal-2025-eurollm9b} versions.
It uses the same BPE-based tokenizer as the previous models, providing broad coverage of European and global languages.
The associated vocabulary contains 128{,}000 units.
The model architecture adopts grouped query attention~\citep{ainslie2023gqa}, pre-layer normalization~\citep{xiong2020layer}, RMS normalization~\citep{zhang2019root}, SwiGLU activation functions~\citep{shazeer2020glu}, and rotary positional embeddings (RoPE;~\citep{su2024roformer}).
The architectural and optimization hyperparameters are summarized in Table~\ref{tab:hyperparameters}.

\subsection{Training Phases}
\label{subsec:training_phases}

\begin{figure}[t]
    \centering
    \includegraphics[width=12cm]{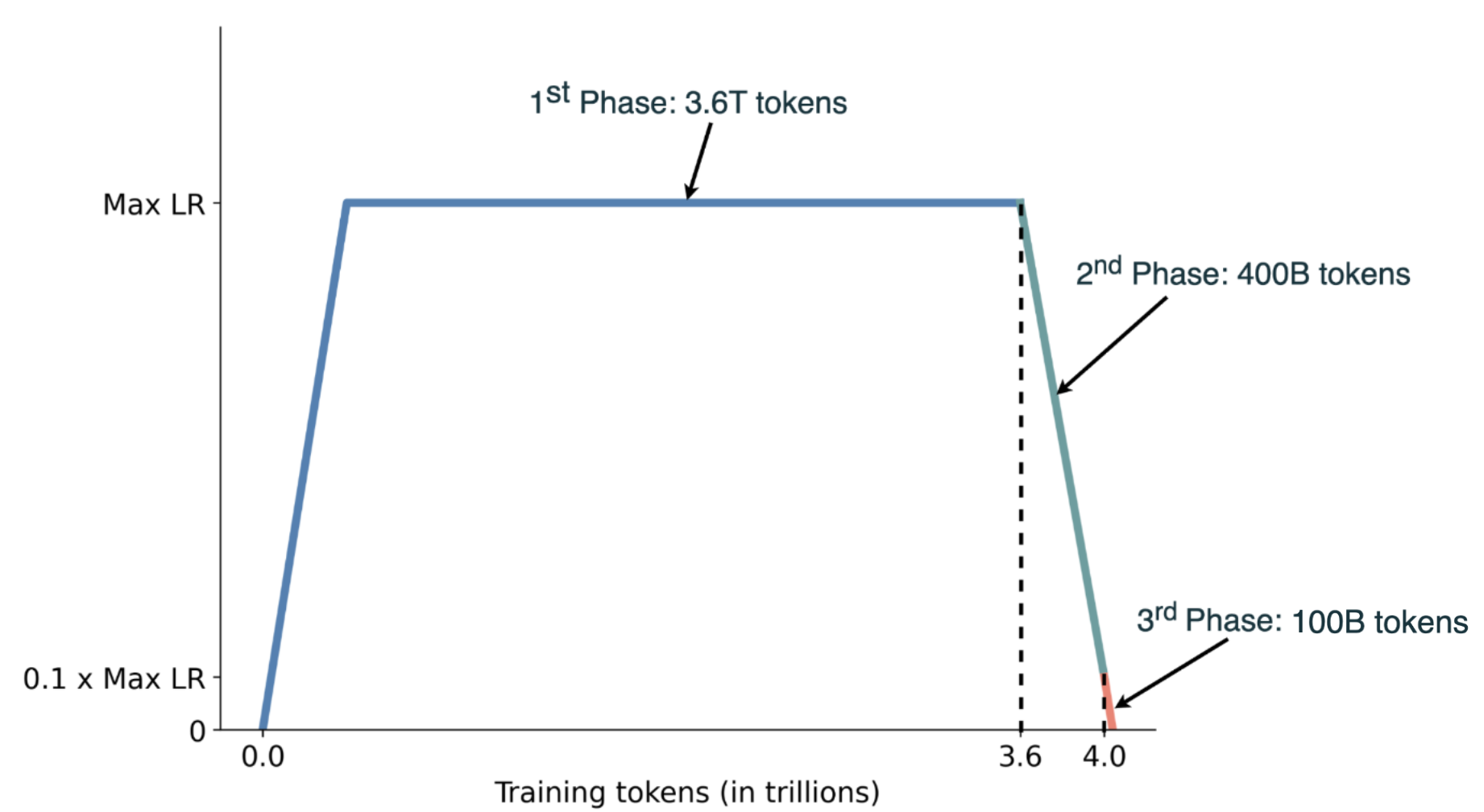}
    \caption{Scheme of the learning rate scheduler.}
    \label{fig:lr_scheduler}
\end{figure}

Similar to the 9B version, \eurollm{} 22B was pretrained with approximately 4T tokens, using a 3-phase training schedule. In the first phase, we train on 3.6T tokens with a 10\% linear warmup to a peak learning rate of $1.5 \times 10^{-4}$, which is kept constant thereafter. We then anneal over 400B tokens, linearly reducing the learning rate to 10\% of its peak, and decay it to zero in the final learning phase. This schedule, illustrated in Figure~\ref{fig:lr_scheduler}, allows us to progressively expose the model to higher quality data~\citep{llama3modelcard}.

Differing from the 9B version, in the final training phase of \eurollm{} 22B, we extend its context window from 4K to 32K, adjusting the maximum sequence length and applying RoPE scaling~\citep{xiong-etal-2024-effective}, increasing the $\theta$ value from $1 \times 10^4$ to $1 \times 10^6$.

\subsection{Dataset}
\label{subsec:dataset}

\begin{table}[t]
\begin{center} 
\footnotesize
\setlength{\tabcolsep}{5ex}
\begin{tabular}{@{\hspace{0.2cm}}lccc}
\toprule
Dataset         & Version   &  \\ \midrule
Europarl \citep{koehn-2005-europarl}      & v8                 &  \\
ParaCrawl \citep{espla-etal-2019-paracrawl}       & v9                  &  \\
MultiParaCrawl \citep{espla-etal-2019-paracrawl}  & v7.1               &  \\
CCMatrix \citep{schwenk2020ccmatrix}       & v1                &  \\
CCAligned \citep{el-kishky-etal-2020-ccaligned}      & v1               &  \\
MultiCCAligned \citep{el-kishky-etal-2020-ccaligned} & v1              &  \\
WikiTitles \citep{tiedemann2012opus}     & v2014           &  \\
WikiMatrix \citep{schwenk2019wikimatrix}     & v1              &  \\
News-Commentary \citep{tiedemann2012opus} & v16                 &  \\
OPUS100 \citep{zhang2020improving} & v1                 &  \\
TildeModel \citep{rozis-skadins-2017-tilde}     & v2018          &  \\
Bible \citep{mayer-cysouw-2014-creating}          & v1               &  \\
Ubuntu \citep{tiedemann2012opus}         & v14.10           &  \\
Tatoeba \citep{tiedemann2012opus}        & v2            &  \\
GNOME \citep{tiedemann2012opus}          & v1          &  \\
GlobalVoices  \citep{tiedemann2012opus}  & v2018q4         &  \\
KDE4  \citep{tiedemann2012opus}          & v2               &  \\
KDE-Doc  \citep{tiedemann2012opus}       & v1             &  \\
PHP \citep{tiedemann2012opus}            & v1           &  \\
Wikipedia \citep{Wo_k_2014}      & v1.0             &  \\
Wikimedia \citep{tiedemann2012opus}      & v20210402       &  \\
JRC \citep{tiedemann2012opus}            & v3.0           &  \\
DGT \citep{tiedemann2012opus}            & v2019              &  \\
EuroPat \citep{europat}        & v3              &  \\
EUbookshop \citep{tiedemann2012opus}     & v2                &  \\
EMEA \citep{tiedemann2012opus}           & v3                &  \\
EUConst \citep{tiedemann2012opus}        & v1               &  \\
tico-19 \citep{anastasopoulos-etal-2020-tico}        & v20201028        &  \\
ECB \citep{tiedemann2012opus}            & v1                 &  \\
Elitr-ECA \citep{williams2021elitr}      & v1            &  \\
MultiUN \citep{eisele-chen-2010-multiun}        & v1               &  \\
OpenOffice \citep{tiedemann2012opus}     & v3                 &  \\
Ada83 \citep{tiedemann2012opus}          & v1                &  \\
infopankki \citep{tiedemann2012opus}     & v1          &  \\
Scielo \citep{soares2018large}         & v1               &  \\
giga-fren \citep{tiedemann2012opus}      & v2               &  \\
UNPC \citep{ziemski-etal-2016-united}           & v1.0           & \\ \bottomrule
\end{tabular} 
\end{center}
\caption{Data sources from which we collect parallel data along with the datasets' version.}
\label{tab:parallel_data_sources}
\end{table}

The pre-training dataset for \eurollm{} 22B builds upon the one used for pre-training \eurollm{} 9B, with a series of targeted modifications aimed at improving overall quality. For completeness, we describe the full dataset below, explicitly highlighting the changes introduced with respect to the 9B setup.

\paragraph{\textbf{English Web Data.}}
For the initial training phase, we use the FineWeb-edu dataset \citep{lozhkov2024fineweb-edu} as the source of our English web data, retaining only documents with an educational score above 2 according to their model-based classifier. In contrast with the 9B training strategy, which the highest-quality FineWeb-edu documents were reserved for the final two stages, we include these documents already in the first phase. The subsequent stages instead sample from the high-quality split of Nemotron-CC \citep{su-etal-2025-nemotron-cc}.

\paragraph{\textbf{Multilingual Web Data.}}
To collect web data for the remaining languages, we employ language-specific strategies based on resource availability.
For high-resource languages~(German, Spanish, French, and Italian), we collect data from RedPajama-Data-v2 \citep{together2023redpajama}, which is pre-deduplicated.  We further apply perplexity filtering using KenLM~\citep{heafield-2011-kenlm}, complemented with a set of heuristic filters. Specifically, we discard documents shorter than 200 characters \citep{mt5-xue2021mt5massivelymultilingualpretrained}, and any page containing the phrase “lorem ipsum,” the word “javascript,” or curly brackets \citep{t5-raffel2023exploringlimitstransferlearning}. Additionally, we remove paragraphs where the fraction of uppercase characters exceeds 40\%, the symbol-to-word ratio is greater than 0.1, or the fraction of words without alphabetic characters exceeds 0.2 \citep{gopher-rae2022scalinglanguagemodelsmethods}.

For the remaining languages, we aggregate data from HPLT \citep{degibert2024}, MADLAD-400 \citep{kudugunta2023}, CulturaX \citep{nguyen2023}, and mC4 \citep{xue2021}. After concatenation, we apply deduplication, language identification, perplexity filtering, and the same set of heuristic filters that we used for the high-resource languages, using a CCNet-based preprocessing pipeline \citep{wenzek2019}.

We classified all our multilingual web data with \href{https://huggingface.co/utter-project/EuroFilter-v1}{\eurofilter{}}~\citep{martins2024eurollm}, our educational filter that assigns a quality score from 0 to 5 to each record.\footnote{The filter is publicly available at \href{https://huggingface.co/utter-project/EuroFilter-v1}{utter-project/EuroFilter-v1}.} This classifier was developed by fine-tuning the mDeBERTa~\citep{he2023debertav} on the quality annotations from the FineWeb-Edu~\citep{lozhkov2024fineweb-edu} classifier, which were translated to all languages supported by \textsc{Tower v2}~\citep{rei-etal-2024-tower}.

Unlike the 9B version, which utilized quality scores only to select data for the final two stages, the 22B version divides all classified web data into three tiers, one for each phase of our training recipe, reserving the highest quality data for the later stages. We publicly release this data as \href{https://huggingface.co/datasets/utter-project/EuroLLM-Multilingual-Data-2512}{EuroWeb}.

\paragraph{\textbf{Parallel Data.}}
Regarding parallel data, we collect sentence-level to-English (xx→en) and from-English (en→xx) parallel data from various public sources, listed in Table~\ref{tab:parallel_data_sources}.

We use Bifixer \citep{prompsit:2020:EAMT} to remove duplicates and ensure translation quality by removing sentence pairs below quality thresholds for Bicleaner \citep{prompsit:2018:WMT,prompsit:2020:EAMT} and \href{https://huggingface.co/Unbabel/wmt22-cometkiwi-da}{CometKiwi-22} \citep{rei-etal-2022-cometkiwi}. For Bicleaner, we use a threshold of 0.6 for Portuguese and of 0.5 for all the other languages, while for \href{https://huggingface.co/Unbabel/wmt22-cometkiwi-da}{CometKiwi-22} we use a threshold of 0.7.

For the second and third training phases, we additionally incorporate document-level parallel data from Europarl \citep{koehn-2005-europarl} and ParaDocs \citep{paradocs}, applying the same filtering criteria.

\paragraph{\textbf{Code / Math Data.}}
We collect code and mathematical data from The Stack \citep{Kocetkov2022TheStack}, the Algebraic-stack \citep{azerbayev2023llemma}, and Open-web-math \citep{paster2023openwebmath}.
For the second and third training phases, we also incorporate the python-edu dataset \citep{benallal2024smollmcorpus} and the training sets of GSM8k \citep{cobbe-etal-2021-gsm8k} and of Mathematics Aptitude Test of Heuristics \citep{hendrycks-etal-2021-math}. In contrast with the previous \eurollm{} versions, we also introduced the FineMath dataset~\citep{allal2025smollm} to improve mathematical reasoning capabilities.

\paragraph{\textbf{Synthetic Math Data.}} For the third training phase, we additionally incorporate approximately 1.7 million samples of synthetic data generated using the Qwen-2.5 models~\citep{qwen2025qwen25technicalreport, yang2024qwen25mathtechnicalreportmathematical}. Starting from the MathInstruct~\citep{toshniwal2024openmathinstruct118millionmath, toshniwal2024openmathinstruct2acceleratingaimath} and MetaMathQA~\citep{yu2024metamathbootstrapmathematicalquestions} datasets, we rewrite the questions and generate new answers using \href{https://huggingface.co/Qwen/Qwen2.5-Math-7B}{Qwen2.5-Math-7B}. The generated answers are then evaluated with LLM-as-a-Judge~\citep{zheng2023judgingllmasajudgemtbenchchatbot}, with \href{https://huggingface.co/Qwen/Qwen2.5-32B-Instruct}{Qwen2.5 32B} acting as the judge, and retaining only samples with a score of at least 9/10.

Additionally, we sample from these datasets to generate multiple-choice questions derived from the original data, using \href{https://huggingface.co/google/gemma-2-9b-it}{Gemma2-9B}. The dataset was further augmented with samples from SlimOrca, which include original prompts and generations from \href{https://huggingface.co/google/gemma-2-9b-it}{Gemma2-9B}, \href{https://huggingface.co/google/gemma-2-27b-it}{Gemma2-27B}~\citep{team2024gemma2}, \href{https://huggingface.co/meta-llama/Llama-3.1-70B-Instruct}{Llama3.1-70B}~\citep{llama3modelcard}, and \href{https://huggingface.co/Qwen/Qwen2.5-32B-Instruct}{Qwen2.5-32B}.
For these answers, \href{https://huggingface.co/Qwen/Qwen2.5-32B-Instruct}{Qwen2.5 32B} provided judgements to ascertain the ``best-of-N'' answer, with ties resolved by randomly selecting one of the top-scoring answers.

\paragraph{\textbf{Higher-quality Data.}}
Regarding high-quality data, we use Wikipedia \citep{wikidump} for all languages and ArXiv \citep{clement2019}, Books \citep{uspdbooks}, and Apollo \citep{wang2024apollo} for English.

For the second and third training phases, we also add the Cosmopedia dataset (second version; \citet{benallal2024smollmcorpus}). 
In the third phase, we further include documents of Cosmopedia translated using Tower \citep{alves2024tower} to German, Spanish, French, Italian, Portuguese, Dutch, Chinese, and Russian. 

\paragraph{\textbf{Long-context data.}}
Supporting longer contexts of up to 32k tokens represents a key distinction from the previous \eurollm{} models. To better accomodate this capability, we incorporated an additional 60B tokens in the final training phase, evenly divided between books and code. This involved upsampling our books corpus and sampling code examples from The Stack v2 \citep{lozhkov2024starcoder2stackv2}, applying a lightweight quality filter, selecting only code examples from repositories with at least 500 stars and 100 forks.

\section{Post Training}
\label{sec:post_training}

We outline the post-training methodology used for \eurollm{}~22B, describing the post-training corpus—released as the new version of \href{https://huggingface.co/datasets/utter-project/EuroBlocks-SFT-2512}{\euroblocks{}} (\S\ref{sec:post_training_data})—and the fine-tuning procedure (\S\ref{sec:supervised-fine-tuning}).

\subsection{Data}
\label{sec:post_training_data}

To construct the new version of \href{https://huggingface.co/datasets/utter-project/EuroBlocks-SFT-2512}{\euroblocks{}}, we build upon the \euroblocks{} series \citep{martins2024eurollm,martins-etal-2025-eurollm9b} by incorporating instructions from additional data sources and responses generated with more capable models.
Following \citet{rei2025towerplus}, we begin with a collection of publicly available datasets \citep{OpenHermes_2.5,dang2024aya,wang2024helpsteer,xu2024magpie}, regenerate answers using multiple open models \citep{deepseekai2025deepseekv3technicalreport,qwen2025qwen25technicalreport,lambert2025tulu,dubey2024llama}, and select the best response using Skywork-Gemma2-27B \citep{liu2024skyworkrewardbagtricksreward} as the reward model. 

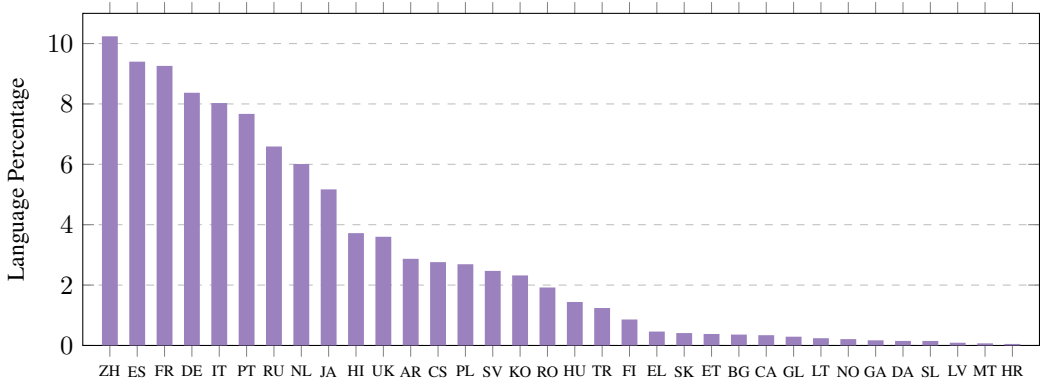
\begin{figure}[!htbp]
\centering
\begin{tikzpicture}
    \begin{axis}[
        ybar=20pt,
        bar width=6pt,
        width=14.3cm, height=6cm,
        enlarge x limits=0.03,
        symbolic x coords={ZH, ES, FR, DE, IT, PT, RU, NL, JA, HI, UK, AR, CS, PL, SV, KO, RO, HU, TR, FI, EL, SK, ET, BG, CA, GL, LT, NO, GA, DA, SL, LV, MT, HR},
        xtick=data,
        xticklabel style={font=\tiny},
        ylabel={Language Percentage},
        ylabel style={font=\footnotesize},
        ymin=0, ymax=11,
        ytick={0, 2, 4, 6, 8, 10},
        ymajorgrids=true,
        xmajorgrids=false,
        grid style=dashed,
        nodes near coords={}, 
        every node near coord/.append style={font=\tiny, /pgf/number format/.cd, fixed, precision=2}
    ]
    \addplot[fill=postTrainingColor, draw=none] coordinates {
        (ZH, 10.24) (ES, 9.40) (FR, 9.26) (DE, 8.37) (IT, 8.03) (PT, 7.67) (RU, 6.59) 
        (NL, 6.01) (JA, 5.17) (HI, 3.72) (UK, 3.60) (AR, 2.87) (CS, 2.76) (PL, 2.69) 
        (SV, 2.47) (KO, 2.32) (RO, 1.92) (HU, 1.44) (TR, 1.24) (FI, 0.86) (EL, 0.46) 
        (SK, 0.41) (ET, 0.38) (BG, 0.36) (CA, 0.34) (GL, 0.29) (LT, 0.24) (NO, 0.21) 
        (GA, 0.17) (DA, 0.15) (SL, 0.15) (LV, 0.09) (MT, 0.07) (HR, 0.05)
    };
    \end{axis}
\end{tikzpicture}
\caption{Language-wise percentage of the post-training corpus, excluding code/math/STEM data. English comprises 60\% of the total data, multilingual content ~20\%, and code/math/STEM data ~20\%.}
\label{fig:language_percentage}
\end{figure}

To broaden domain coverage, we further augment the data with \href{https://huggingface.co/datasets/NousResearch/Hermes-3-Dataset}{Hermes-3} \citep{teknium2024hermes3technicalreport}, \href{https://huggingface.co/datasets/allenai/tulu-3-sft-mixture}{T{\"u}lu~3} \citep{lambert2025tulu}, and \href{https://huggingface.co/datasets/nvidia/Nemotron-Post-Training-Dataset-v2}{Nemotron~V2} \citep{NemotronPostTrainingDatasetV2}. We also include two million STEM-oriented\footnote{For: Science, Technology, Engineering and Mathematics.} samples from \href{https://huggingface.co/datasets/nvidia/Nemotron-Post-Training-Dataset-v1}{Nemotron~V1} \citep{NemotronPostTrainingDatasetV1}. These sources provide diverse prompts and responses spanning general conversation, coding, mathematical problem solving, and other STEM content.
Many collected samples contained structured reasoning traces. We remove all such traces, yielding a fully non-reasoning instruction–response corpus. We then perform instruction-level deduplication and discard poorly formatted samples.
The resulting dataset contains approximately 10.6 million multilingual examples (see Figure~\ref{fig:language_percentage} for the language distribution).

\subsection{Supervised fine-tuning}
\label{sec:supervised-fine-tuning}

To obtain \eurollm{}-22B-Instruct, our instruction-following model, we fine-tune our base model on \euroblocks{}-22B using a maximum context length of $32,768$ tokens. 
Training optimizes the standard cross-entropy objective, computing the loss only on the target tokens.
We train for $5$ epochs using \texttt{bfloat16} mixed precision, sequence packing, and a cosine learning rate scheduler with a maximum learning rate of $1\times 10^{-5}$ and $125$ warmup steps.

We adopt Axolotl\footnote{\url{https://axolotl.ai/}} coupled with Liger-Kernel\footnote{\url{https://github.com/linkedin/Liger-Kernel}} \citep{hsu2025ligerkernel}, which significantly improves training efficiency and reduces memory consumption.
We enable optimized implementations from Liger-Kernel for RoPE, RMSNorm, GLU activation, layer normalization, and fused linear cross-entropy.
Complete training configurations---including the Axolotl YAML configuration---are available in the model card accompanying each released \eurollm{} model.

\section{Evaluation}
\label{sec:evaluation}

Our evaluations span a broad set of benchmarks commonly used for instruction-tuned models, covering both English and multilingual settings. 
The English suite includes instruction-following, general-knowledge, and STEM tasks, while the multilingual suite covers general-knowledge, STEM, and translation tasks.
We release our evaluation framework to ensure reproducibility and facilitate future research.\footnote{\url{https://github.com/deep-spin/eurollm-eval}.}

\subsection{English Benchmarks}

\paragraph{Instruction-following.}
We evaluate instruction following using IFEval \citep{kovalevskyi-etal-2024-ifeval}, a suite of prompts designed to assess a model’s ability to follow explicit instructions (e.g., avoiding a specific word in the answer or structuring the response into a given number of sections).

\paragraph{General knowledge.} We employ several benchmarks, including Hellaswag \citep{zellers-etal-2019-hellaswag}, MMLU \citep{hendrycks-etal-2021-mmlu}, MMLU-Pro \citep{wang-etal-2024-mmlupro}, and BBH \citep{suzgun-etal-2023-challenging}, which together assess commonsense reasoning, broad knowledge, and multitask generalization.

\paragraph{STEM.} We evaluate STEM knowledge using several benchmarks, including ARC-C \citep{clark-etal-2018-arc}, the challenge split of the ARC multiple-choice science exam corpus, and GPQA $\blacklozenge$ \citep{rein2024gpqa}, a set of difficult graduate-level physics problems. 
For mathematics, we use GSM8K \citep{cobbe-etal-2021-gsm8k}, which contains grade-school math word problems requiring multi-step reasoning, and MATH-500 \citep{lightman2023let}, which includes high-school and early undergraduate math problems.
For coding, we use HumanEval \citep{chen2021evaluating}, a benchmark for generating python code from natural language descriptions.

\subsection{Multilingual Benchmarks}

\paragraph{General knowledge.} We evaluate multilingual general knowledge using multilingual Hellaswag, MMMLU, and MMLU-ProX \citep{dac2023okapi,xuan2025mmluprox}, which are multilingual extensions of the Hellaswag and MMLU benchmarks, and a multilingual adaptation of MMLU-Pro, respectively.

\paragraph{STEM.} We evaluate multilingual STEM knowledge using multilingual ARC-C \citep{dac2023okapi} and MGSM \citep{shi2022language}, which are a multilingual extension of the ARC-C benchmark and a manually translated subset of 250 GSM8K questions into 10 languages, respectively.

\paragraph{Translation.} We evaluate machine translation using FLORES-200 \citep{nllb2024}, a benchmark for translation between English and low-resource languages. We also employ WMT24++ \citep{deutsch-etal-2025-wmt24}, an extension of WMT24 \citep{kocmi2024findings} covering 55 languages and dialects, and WMT25 \citep{kocmi2025findings}, the latest WMT benchmark for translation across diverse language pairs.

\paragraph{Multilingual coverage.} All multilingual benchmarks are restricted to the languages supported by EuroLLM-22B, which include Bulgarian, Croatian, Czech, Danish, Dutch, English, Estonian, Finnish, French, German, Greek, Hungarian, Irish, Italian, Latvian, Lithuanian, Maltese, Polish, Portuguese, Romanian, Slovak, Slovenian, Spanish, Swedish, Arabic, Catalan, Chinese, Galician, Hindi, Japanese, Korean, Norwegian, Russian, Turkish, and Ukrainian.

\subsection{Baselines}

We compare \href{https://huggingface.co/utter-project/EuroLLM-22B-Instruct-2512}{EuroLLM-22B} and our newly released \href{https://huggingface.co/utter-project/EuroLLM-9B-Instruct-2512}{EuroLLM-9B} with instruction-tuned baselines of comparable size, including both European and non-European models, and encompassing fully open as well as open-weights models.

\paragraph{European models.} We compare with the fully open European baselines \href{https://huggingface.co/swiss-ai/Apertus-8B-Instruct-2509}{Apertus-8B} and \href{https://huggingface.co/swiss-ai/Apertus-70B-Instruct-2509}{Apertus-70B}~\citep{hernándezcano2025apertus}. 
We also include an open-weights baseline, \href{https://huggingface.co/mistralai/Mistral-Small-3.2-24B-Instruct-2506}{Mistral-3.2-24B}~\citep{jiang2023mistral}. 
Additionally, for completeness and historical comparison, we separately compare against our previous EuroLLM models, \href{https://huggingface.co/utter-project/EuroLLM-9B-Instruct}{EuroLLM-9B \textit{(old)}} and \href{https://huggingface.co/utter-project/EuroLLM-22B-Instruct-Preview}{EuroLLM-22B-Preview}~\citep{martins-etal-2025-eurollm9b}.

\paragraph{Non-European models.} The fully open baselines include \href{https://huggingface.co/allenai/Olmo-3-7B-Instruct}{OLMo-3-7B} and \href{https://huggingface.co/allenai/Olmo-3.1-32B-Instruct}{OLMo-3.1-32B} \citep{olmo2025olmo}.
We additionally compare with open-weights baselines such as \href{https://huggingface.co/meta-llama/Llama-3.1-8B-Instruct}{Llama-3.1-8B}, \href{https://huggingface.co/meta-llama/Llama-3.3-70B-Instruct}{Llama-3.3-70B} \citep{dubey2024llama}, \href{https://huggingface.co/google/gemma-3-12b-it}{Gemma-3-12B}, \href{https://huggingface.co/google/gemma-3-27b-it}{Gemma-3-27B} \citep{team2025gemma}, \href{https://huggingface.co/Qwen/Qwen3-14B}{Qwen-3-14B}, \href{https://huggingface.co/Qwen/Qwen3-32B}{Qwen-3-32B}, and \href{https://huggingface.co/Qwen/Qwen3-30B-A3B}{Qwen-3-30B-A3B} \citep{yang2025qwen3}.

\subsection{Evaluation Protocol}

\paragraph{Inference parameters.}

To ensure a fair comparison between models, we use the generation parameters recommended by the authors when available and otherwise default to greedy decoding, performing all generation in non-reasoning mode. Accordingly, inference for Qwen-3 is performed with a temperature of 0.7, top-p of 0.8, top-k of 20, min-p of 0, and a presence penalty of 1.5, as suggested by \citet{yang2025qwen3}. Additionally, all models are allowed to generate up to their maximum length, giving more verbose models the full opportunity to produce their outputs.

\paragraph{Answer assessment.}

All tasks are evaluated using LLM-as-a-judge. For non-translation tasks, this approach primarily avoids the limitations of rule-based extraction, which can be unreliable for some models that sometimes fail to format their outputs correctly.\footnote{We show in \autoref{apd:regex_vs_llm} that LLM-as-a-judge achieves substantially higher correlation with human judgments than rule-based extraction.} Specifically, a high-capacity judge is provided with the question, the generated answer, and the ground truth, and is asked to determine whether the generated answer is equivalent to the ground truth.\footnote{We provide the full assessment prompts used in \autoref{sec:assessment_prompt}.} As judges, we use \href{https://huggingface.co/nvidia/Llama-3_3-Nemotron-Super-49B-v1_5}{Nemotron-49B} \citep{bercovich2025llamanemotronefficientreasoningmodels}, \href{https://huggingface.co/openai/gpt-oss-120b}{GPT-OSS-120B} \citep{openai2025gptoss120bgptoss20bmodel}, and \href{https://huggingface.co/Qwen/Qwen3-235B-A22B-Instruct-2507}{Qwen3-235B-A22B} \citep{yang2025qwen3}, and aggregate their judgments by mean. 
For translation, we use \href{https://huggingface.co/Unbabel/wmt22-comet-da}{COMET-22} \citep{rei-etal-2022-comet}, providing the source, generated translation, and gold reference for scoring.

\subsection{Results}

This section documents performance results on English benchmarks (\autoref{tab:results_en}) and aggregate results on multilingual benchmarks restricted to European languages (\autoref{tab:results_eu}). Aggregate results over all multilingual benchmarks (\autoref{tab:results_xx}) and over non-EU languages (\autoref{tab:results_non_eu}), as well as detailed per-language and per-language-pair results, are provided in \autoref{sec:detailed_results}.

\begin{table}[!htbp]
\setlength{\tabcolsep}{1.72pt}
\small
\begin{tabular}{lcccccccccc}
\toprule
\multicolumn{1}{c}{} & \multicolumn{1}{c}{\textbf{IF}} & \multicolumn{4}{c}{\textbf{General}} & \multicolumn{5}{c}{\textbf{STEM}} \\
\cmidrule(lr){2-2} \cmidrule(lr){3-6} \cmidrule(lr){7-11}
\textbf{Model} & \textbf{IFEval} & \textbf{Hellaswag} & \textbf{MMLU} & \textbf{\makecell{MMLU\\Pro}} & \textbf{BBH} & \textbf{ARC-C} & \textbf{GPQA$\blacklozenge$} & \textbf{GSM8K} & \textbf{\makecell{MATH\\500}} & \textbf{\makecell{Human\\Eval}} \\
\midrule
\multicolumn{11}{c}{\textbf{\textit{Fully-open}}} \\
\midrule
\multicolumn{11}{l}{\textbf{\textit{European}}} \\
\addlinespace[2pt]
EuroLLM-9B & 62.4 & 53.0 & 65.5 & 42.3 & 45.8 & 85.9 & 21.0 & 74.6 & 36.9 & 50.8 \\
EuroLLM-22B & \underline{67.2} & 69.7 & \underline{69.8} & \underline{50.8} & 55.3 & \underline{89.8} & \underline{26.8} & \underline{85.5} & \underline{54.5} & \underline{53.9} \\
\addlinespace[3pt]
Apertus-8B & 59.1 & 58.1 & 57.3 & 32.7 & 42.8 & 75.5 & 24.6 & 67.7 & 26.9 & 39.0 \\
Apertus-70B & 61.2 & \underline{74.6} & 67.9 & 41.9 & \underline{56.1} & 84.7 & 21.4 & 80.0 & 42.3 & 44.5 \\
\addlinespace[4pt]
\multicolumn{11}{l}{\textbf{\textit{Non-European}}} \\
\addlinespace[2pt]
OLMo-3-7B & 75.5 & 42.8 & 69.3 & 56.9 & 75.5 & 86.1 & 33.2 & 93.4 & 84.2 & 86.4 \\
OLMo-3.1-32B & \textbf{84.2} & \textbf{75.8} & \textbf{80.1} & \textbf{66.5} & \textbf{85.3} & \textbf{93.6} & \textbf{36.0} & \textbf{94.5} & \textbf{85.7} & \textbf{87.6} \\
\midrule
\multicolumn{11}{c}{\textbf{\textit{Open-weights}}} \\
\midrule
\multicolumn{11}{l}{\textbf{\textit{European}}} \\
\addlinespace[2pt]
Mistral-3.2-24B & 65.7 & 84.0 & 77.3 & 67.4 & 78.1 & 93.4 & 47.5 & 95.5 & 81.5 & 73.6 \\
\addlinespace[4pt]
\multicolumn{11}{l}{\textbf{\textit{Non-European}}} \\
\addlinespace[2pt]
Llama-3.1-8B & 63.8 & 44.0 & 68.3 & 45.8 & 57.6 & 84.3 & 26.8 & 84.9 & 49.4 & 59.3 \\
Llama-3.3-70B & 82.8 & 86.3 & 84.6 & 70.4 & 82.3 & 94.5 & 46.6 & \textbf{96.4} & 74.6 & 71.1 \\
\addlinespace[3pt]
Gemma-3-12B & 76.5 & 83.2 & 76.1 & 59.9 & 78.4 & 92.3 & 37.2 & 95.0 & 85.3 & 69.1 \\
Gemma-3-27B & 80.7 & 84.5 & 80.4 & 66.6 & 82.2 & 93.5 & 47.6 & 96.0 & 88.5 & 73.2 \\
\addlinespace[3pt]
Qwen-3-14B & 81.6 & 86.7 & 81.2 & 71.1 & 83.5 & 94.3 & 56.6 & 95.0 & 86.9 & 74.6 \\
Qwen-3-32B & 81.9 & 87.4 & 84.0 & 74.1 & 83.7 & 95.2 & 54.7 & 95.2 & 85.7 & \textbf{75.0} \\
Qwen-3-30B-A3B & \textbf{83.7} & \textbf{88.2} & \textbf{85.0} & \textbf{76.7} & \textbf{86.1} & \textbf{96.0} & \textbf{58.6} & 96.3 & \textbf{89.7} & \textbf{75.0} \\
\bottomrule
\end{tabular}
\caption{Results on English benchmarks. \textbf{Bold} indicates the best score per benchmark within each section (\textit{Fully-open} or \textit{Open-weights}). \underline{Underlined} indicates the best \textit{Fully-open} European score.}
\label{tab:results_en}
\end{table}

\begin{table}[!htbp]
\setlength{\tabcolsep}{1.74pt}
\small
\begin{tabular}{lcccccccc}
\toprule
\multicolumn{1}{c}{} & \multicolumn{3}{c}{\textbf{General}} & \multicolumn{2}{c}{\textbf{STEM}} & \multicolumn{3}{c}{\textbf{Translation}} \\
\cmidrule(lr){2-4} \cmidrule(lr){5-6} \cmidrule(lr){7-9}
\textbf{Model} & \textbf{Hellaswag} & \textbf{MMMLU} & \textbf{MMLU-ProX} & \textbf{ARC-C} & \textbf{MGSM} & \textbf{FLORES} & \textbf{WMT24++} & \textbf{WMT25} \\
\midrule
\multicolumn{9}{c}{\textbf{\textit{Fully-open}}} \\
\midrule
\multicolumn{9}{l}{\textbf{\textit{European}}} \\
\addlinespace[2pt]
EuroLLM-9B & 49.9 & 61.5 & 39.0 & 80.7 & 71.0 & \textbf{\underline{88.9}} & 83.6 & 80.4 \\
EuroLLM-22B & 62.6 & \underline{65.6} & \underline{46.8} & \textbf{\underline{84.1}} & \underline{77.8} & \textbf{\underline{88.9}} & \textbf{\underline{83.9}} & 80.9 \\
\addlinespace[3pt]
Apertus-8B & 50.9 & 54.0 & 30.4 & 71.0 & 61.4 & 87.8 & 81.5 & 80.0 \\
Apertus-70B & \textbf{\underline{68.6}} & 61.7 & 37.8 & 79.6 & 73.6 & 85.1 & 76.0 & \textbf{\underline{82.0}} \\
\addlinespace[4pt]
\multicolumn{9}{l}{\textbf{\textit{Non-European}}} \\
\addlinespace[2pt]
OLMo-3-7B & 30.0 & 49.3 & 43.0 & 54.5 & 80.6 & 68.0 & 62.4 & 40.3 \\
OLMo-3.1-32B & 49.2 & \textbf{68.2} & \textbf{58.9} & 79.8 & \textbf{88.8} & 80.1 & 74.3 & 57.2 \\
\midrule
\multicolumn{9}{c}{\textbf{\textit{Open-weights}}} \\
\midrule
\multicolumn{9}{l}{\textbf{\textit{European}}} \\
\addlinespace[2pt]
Mistral-3.2-24B & \textbf{84.3} & 76.0 & 65.6 & 90.0 & 90.8 & 86.7 & 79.7 & 70.2 \\
\addlinespace[4pt]
\multicolumn{9}{l}{\textbf{\textit{Non-European}}} \\
\addlinespace[2pt]
Llama-3.1-8B & 37.7 & 54.3 & 35.6 & 69.0 & 75.6 & 83.6 & 75.1 & 68.9 \\
Llama-3.3-70B & 74.7 & 79.9 & 68.0 & 91.1 & \textbf{93.0} & 88.0 & 82.2 & 77.2 \\
\addlinespace[3pt]
Gemma-3-12B & 74.5 & 70.3 & 54.9 & 87.9 & 87.5 & 88.0 & 83.2 & 82.4 \\
Gemma-3-27B & 76.4 & 75.8 & 61.6 & 90.8 & 89.9 & \textbf{88.8} & \textbf{84.0} & \textbf{83.9} \\
\addlinespace[3pt]
Qwen-3-14B & 77.5 & 75.8 & 67.5 & 90.5 & 90.3 & 85.6 & 81.4 & 74.9 \\
Qwen-3-32B & 80.5 & 79.9 & 71.3 & \textbf{93.1} & 92.0 & 86.0 & 81.8 & 75.9 \\
Qwen-3-30B-A3B & 79.3 & \textbf{80.6} & \textbf{73.1} & \textbf{93.1} & 91.4 & 86.3 & 82.2 & 77.9 \\
\bottomrule
\end{tabular}
\caption{Results on multilingual benchmarks restricted to the 24 official European Union languages. \textbf{Bold} indicates the best score per benchmark within each section (\textit{Fully-open} or \textit{Open-weights}). \underline{Underlined} indicates the best \textit{Fully-open} European score.}
\label{tab:results_eu}
\end{table}

\paragraph{Results for post-trained models.}
The instruction-tuned results are summarized in Tables~\ref{tab:results_en} and \ref{tab:results_eu}, with full per-language breakdowns reported in \autoref{sec:detailed_results}.
Across the benchmark suite, the new \eurollm{}-9B consistently improves over Apertus-8B, while \eurollm{}-22B is the strongest model among the fully open European systems considered, confirming a clear scaling trend within the \eurollm{} family.
A particularly informative comparison is against Apertus-70B. Here, \eurollm{}-22B operates with roughly one third of the parameters, yet is frequently competitive and in several settings achieves higher scores across both English and multilingual European evaluations, indicating that its instruction tuning and multilingual design translate into robust downstream behavior rather than gains concentrated in a narrow subset of tasks.
Taken together, while the \eurollm{} family still trails the very best open-weights models overall, it offers the strongest fully open European alternative as to date.

\paragraph{Results for pre-trained models.}
The base-model results are reported in \autoref{sec:base_model_evaluation}.
\eurollm{}-22B-Base shows consistent gains over \eurollm{}-9B-Base, aligning with the expected benefits from scaling while remaining broadly competitive with the strongest fully-open European baselines. The remaining gap to Apertus-70B-Base should be interpreted in the context of substantially different training regimes, as \eurollm{}-22B is trained on approximately 4T tokens, whereas Apertus-70B reports pre-training on 15T tokens at a much larger parameter scale.
These results suggest that \eurollm{} achieves strong quality with a comparatively modest token budget, and that increasing the amount of high-quality training data is a promising direction for further closing the gap.

\subsection{Post-training Analysis and Discussion}

To isolate the effect of our updated post-training recipe, \autoref{tab:results_eurollm_improvements_en} and \autoref{tab:results_eurollm_improvements_eu} compare the previous (\textit{old}) and current (\textit{new}) instruction-tuned \eurollm{} checkpoints (9B and 22B) on identical English and multilingual evaluation suites, with the multilingual suite restricted to European languages. Additional results by language and language pair are provided in \autoref{sec:detailed_results}.

\begin{table}[!htbp]
\centering
\setlength{\tabcolsep}{3pt}
\small
\begin{tabular}{lcccccccccc}
\toprule
\multicolumn{1}{c}{} & \multicolumn{1}{c}{\textbf{IF}} & \multicolumn{4}{c}{\textbf{General}} & \multicolumn{5}{c}{\textbf{STEM}} \\
\cmidrule(lr){2-2} \cmidrule(lr){3-6} \cmidrule(lr){7-11}
\textbf{Model} & \textbf{IFEval} & \textbf{Hellaswag} & \textbf{MMLU} & \textbf{\makecell{MMLU\\Pro}} & \textbf{BBH} & \textbf{ARC-C} & \textbf{GPQA$\blacklozenge$} & \textbf{GSM8K} & \textbf{\makecell{MATH\\500}} & \textbf{\makecell{Human\\Eval}} \\
\midrule
9B \textit{(old)} & 46.3 & 47.2 & 57.5 & 31.4 & 41.2 & 76.2 & 17.3 & 69.3 & 36.7 & 35.4 \\
9B \textit{(new)} & \textbf{62.4} & \textbf{53.0} & \textbf{65.5} & \textbf{42.3} & \textbf{45.8} & \textbf{85.9} & \textbf{21.0} & \textbf{74.6} & \textbf{36.9} & \textbf{50.8} \\
\midrule
22B \textit{(old)} & 61.6 & \textbf{74.3} & 65.3 & 43.0 & 53.9 & 85.6 & 25.1 & 82.8 & 48.6 & 43.1 \\
22B \textit{(new)} & \textbf{67.2} & 69.7 & \textbf{69.8} & \textbf{50.8} & \textbf{55.3} & \textbf{89.8} & \textbf{26.8} & \textbf{85.5} & \textbf{54.5} & \textbf{53.9} \\
\bottomrule
\end{tabular}
\caption{Improvements on English benchmarks achieved from the previous versions of EuroLLM.}
\label{tab:results_eurollm_improvements_en}
\end{table}

\begin{table}[!htbp]
\centering
\setlength{\tabcolsep}{3pt}
\small
\begin{tabular}{lcccccccc}
\toprule
\multicolumn{1}{c}{} & \multicolumn{3}{c}{\textbf{General}} & \multicolumn{2}{c}{\textbf{STEM}} & \multicolumn{3}{c}{\textbf{Translation}} \\
\cmidrule(lr){2-4} \cmidrule(lr){5-6} \cmidrule(lr){7-9}
\textbf{Model} & \textbf{Hellaswag} & \textbf{MMMLU} & \textbf{MMLU-ProX} & \textbf{ARC-C} & \textbf{MGSM} & \textbf{FLORES} & \textbf{WMT24++} & \textbf{WMT25} \\
\midrule
9B \textit{(old)} & \textbf{55.5} & 55.0 & 30.1 & 73.5 & 61.9 & 88.8 & 83.5 & * \\
9B \textit{(new)} & 49.9 & \textbf{61.5} & \textbf{39.0} & \textbf{80.7} & \textbf{71.0} & \textbf{88.9} & \textbf{83.6} & \textbf{80.4} \\
\midrule
22B \textit{(old)} & \textbf{66.4} & 61.2 & 39.3 & 80.0 & 73.9 & \textbf{88.9} & \textbf{83.9} & * \\
22B \textit{(new)} & 62.6 & \textbf{65.6} & \textbf{46.8} & \textbf{84.1} & \textbf{77.8} & \textbf{88.9} & \textbf{83.9} & \textbf{80.9} \\
\bottomrule
\end{tabular}
\caption{Improvements on multilingual benchmarks, restricted to the 24 official European Union languages, relative to previous versions of \eurollm{}. \textit{*Not evaluated because the required context exceeds the model's maximum context length.}}
\label{tab:results_eurollm_improvements_eu}
\end{table}

\paragraph{Result Analysis and Discussion.}
Across both English and multilingual evaluations, the new \eurollm{} checkpoints show consistent improvements, with the largest gains in instruction following and in knowledge- and STEM-focused problem solving (including coding).
These gains come with translation quality remaining essentially unchanged, suggesting that the updated post-training recipe strengthens general assistant behavior and multilingual reasoning without meaningful trade-offs in translation.
The longer maximum context length also closes prior evaluation gaps and enables coverage of additional long-context benchmarks (e.g., WMT25).
Overall, the results show that the improved post-training recipe yields a significant performance gap over the previous \eurollm{} checkpoints, even though both versions start from similar base models trained on a comparatively modest pre-training budget of 4T tokens.

\section{Conclusions}
\label{sec:conclusions}

In this work, we present \eurollm{}-22B, detailing its development from data collection and filtering to pre-training and post-training procedures. 
We release both the base and instruction-tuned variants of \eurollm{}-22B, accompanied by extensive evaluations on multilingual general benchmarks and machine translation tasks. Alongside the 22B models, we release improved versions of our 9B models, incorporating long-context extension and our improved post-training. To further support research and downstream applications, we also release the new \euroblocks{} dataset, a multilingual instruction dataset designed to improve the model’s performance across European languages; EuroWeb, our multilingual pretraining data; and our pre-training and evaluation codebases.
Collectively, these resources contribute to advancing multilingual language modeling and provide a foundation for future research in European language understanding and generation.

\subsubsection*{Acknowledgments}

Part of this work was supported by the EU’s Horizon Europe Research and Innovation Actions (UTTER, contract 101070631), by the project DECOLLAGE (ERC-2022-CoG 101088763), and by the Portuguese Recovery and Resilience Plan through project C645008882-00000055 (Center for Responsible AI). We thank EuroHPC for the HPC resources used to support this work through grant EHPC-EXT-2023E01-042 and grants EHPC-AI-2024A01-085 and EHPC-AI-2024A05-044.

\bibliography{EuroLLM}
\bibliographystyle{iclr2025_conference}

\newpage
\appendix
\section{Detailed Results for Instruction-Tuned Models}
\label{sec:detailed_results}

This appendix summarizes aggregate results across all multilingual benchmarks, including a subset limited to non-EU languages. We also provide benchmark-level results, broken down by language.

\begin{table}[!h]
\setlength{\tabcolsep}{1.7pt}
\small
\resizebox{\textwidth}{!}{
\begin{tabular}{lcccccccc}
\toprule
\multicolumn{1}{c}{} & \multicolumn{3}{c}{\textbf{General}} & \multicolumn{2}{c}{\textbf{STEM}} & \multicolumn{3}{c}{\textbf{Translation}} \\
\cmidrule(lr){2-4} \cmidrule(lr){5-6} \cmidrule(lr){7-9}
\textbf{Model} & \textbf{Hellaswag} & \textbf{MMMLU} & \textbf{MMLU-ProX} & \textbf{ARC-C} & \textbf{MGSM} & \textbf{FLORES} & \textbf{WMT24++} & \textbf{WMT25} \\
\midrule
\multicolumn{9}{c}{\textbf{\textit{Fully-open}}} \\
\midrule
\multicolumn{9}{l}{\textbf{\textit{European}}} \\
\addlinespace[3pt]
EuroLLM-9B \textit{(old)} & 53.9 & 53.8 & 29.0 & 72.2 & 60.5 & 88.8 & 83.2 & * \\
EuroLLM-9B \textit{(new)} & 49.1 & 60.2 & 37.7 & 79.6 & 67.3 & 88.8 & 83.3 & 80.2 \\
EuroLLM-22B \textit{(old)} & 65.0 & 59.7 & 37.9 & 78.7 & 71.9 & \textbf{\underline{88.9}} & \textbf{\underline{83.6}} & * \\
EuroLLM-22B \textit{(new)} & 62.3 & \underline{64.1} & \underline{45.3} & \textbf{\underline{82.7}} & \underline{76.1} & 88.8 & 83.5 & 79.3 \\
\bottomrule
\end{tabular}}
\caption{Results on multilingual benchmarks. \textbf{Bold} indicates the best score per benchmark within each section (\textit{Fully-open} or \textit{Open-weights}). \underline{Underlined} indicates the best \textit{Fully-open} European score. \textit{*Not evaluated because the required context exceeds the model's maximum context length.}}
\label{tab:results_xx}
\end{table}

\begin{table}[!h]
\setlength{\tabcolsep}{1.5pt}
\small
\resizebox{\textwidth}{!}{
\begin{tabular}{lcccccccc}
\toprule
\multicolumn{1}{c}{} & \multicolumn{3}{c}{\textbf{General}} & \multicolumn{2}{c}{\textbf{STEM}} & \multicolumn{3}{c}{\textbf{Translation}} \\
\cmidrule(lr){2-4} \cmidrule(lr){5-6} \cmidrule(lr){7-9}
\textbf{Model} & \textbf{Hellaswag} & \textbf{MMMLU} & \textbf{MMLU-ProX} & \textbf{ARC-C} & \textbf{MGSM} & \textbf{FLORES} & \textbf{WMT24++} & \textbf{WMT25} \\
\midrule
\multicolumn{9}{c}{\textbf{\textit{Fully-open}}} \\
\midrule
\multicolumn{9}{l}{\textbf{\textit{European}}} \\
\addlinespace[2pt]
EuroLLM-9B \textit{(old)} & 50.1 & 51.3 & 27.9 & 69.5 & 59.2 & 88.6 & 82.5 & * \\
EuroLLM-9B \textit{(new)} & 47.3 & 57.7 & 36.5 & 77.2 & 63.6 & \textbf{\underline{88.7}} & 82.7 & 80.2 \\
EuroLLM-22B \textit{(old)} & 61.4 & 56.6 & 36.4 & 76.1 & 69.9 & \textbf{\underline{88.7}} & \textbf{\underline{82.8}} & * \\
EuroLLM-22B \textit{(new)} & 61.7 & \underline{61.1} & \underline{43.8} & \textbf{\underline{79.8}} & \underline{74.4} & \textbf{\underline{88.7}} & \textbf{\underline{82.8}} & 78.7 \\
\bottomrule
\end{tabular}}

\caption{Results on multilingual benchmarks restricted to non-EU languages. \textbf{Bold} indicates the best score per benchmark within each section (\textit{Fully-open} or \textit{Open-weights}). \underline{Underlined} indicates the best \textit{Fully-open} European score. \textit{*Not evaluated because the required context exceeds the model's maximum context length.}}
\label{tab:results_non_eu}
\end{table}

\begin{table}[!h]
\setlength{\tabcolsep}{1.7pt}
\small
\resizebox{\textwidth}{!}{
\begin{tabular}{lccccccccccccccccc}
\toprule
\multicolumn{1}{c}{} & \multicolumn{12}{c}{\textbf{EU}} & \multicolumn{5}{c}{\textbf{Non-EU}} \\
\cmidrule(lr){2-13} \cmidrule(lr){14-18}
\textbf{Model} & \textbf{da} & \textbf{de} & \textbf{es} & \textbf{fr} & \textbf{hr} & \textbf{hu} & \textbf{it} & \textbf{nl} & \textbf{pt} & \textbf{ro} & \textbf{sk} & \textbf{sv} & \textbf{ar} & \textbf{ca} & \textbf{hi} & \textbf{ru} & \textbf{uk} \\
\midrule
\multicolumn{18}{c}{\textbf{\textit{Fully-open}}} \\
\midrule
\multicolumn{18}{l}{\textbf{\textit{European}}} \\
\addlinespace[2pt]
EuroLLM-9B \textit{(old)} & 57.4 & 57.2 & 57.9 & 57.4 & 53.4 & 52.3 & 55.7 & 56.5 & 55.7 & 52.0 & 52.6 & 57.5 & 50.6 & 49.7 & 44.3 & 54.6 & 51.3 \\
EuroLLM-9B \textit{(new)} & 48.9 & 47.4 & 52.4 & 53.1 & 48.1 & 48.4 & 49.6 & 53.4 & 49.4 & 52.6 & 44.9 & 51.0 & 45.8 & 50.2 & 44.9 & 50.0 & 46.1 \\
EuroLLM-22B \textit{(old)} & 68.2 & 68.3 & 68.5 & 69.4 & 61.9 & 60.6 & 67.0 & 68.7 & 68.4 & 65.7 & 62.9 & 68.0 & 59.3 & 64.4 & 55.2 & 65.3 & 63.0 \\
EuroLLM-22B \textit{(new)} & 60.1 & 64.3 & 67.2 & 65.6 & 59.1 & 58.8 & 64.5 & 65.8 & 65.5 & 59.4 & 56.5 & 63.8 & 63.1 & 61.5 & 56.0 & 65.4 & 62.3 \\
\bottomrule
\end{tabular}}
\caption{Per-language performance on multilingual Hellaswag. \textbf{Bold} indicates the best score per benchmark within each section (\textit{Fully-open} or \textit{Open-weights}). \underline{Underlined} indicates the best \textit{Fully-open} European score.}
\label{tab:results_mhellaswag}
\end{table}

\begin{table}[!htbp]
\setlength{\tabcolsep}{3pt}
\resizebox{\textwidth}{!}{
\begin{tabular}{lcccccccccccccccccc}
\toprule
\multicolumn{1}{c}{} & \multicolumn{12}{c}{\textbf{EU}} & \multicolumn{6}{c}{\textbf{Non-EU}} \\
\cmidrule(lr){2-13} \cmidrule(lr){14-19}
\textbf{Model} & \textbf{da} & \textbf{de} & \textbf{es} & \textbf{fr} & \textbf{hr} & \textbf{hu} & \textbf{it} & \textbf{nl} & \textbf{pt} & \textbf{ro} & \textbf{sk} & \textbf{sv} & \textbf{ar} & \textbf{ca} & \textbf{hi} & \textbf{ru} & \textbf{uk} & \textbf{zh} \\
\midrule
\multicolumn{19}{c}{\textbf{\textit{Fully-open}}} \\
\midrule
\multicolumn{19}{l}{\textbf{\textit{European}}} \\
\addlinespace[2pt]
EuroLLM-9B \textit{(old)} & 55.2 & 54.9 & 56.6 & 55.8 & 52.7 & 52.8 & 55.6 & 55.7 & 56.1 & 55.3 & 54.0 & 54.7 & 49.0 & 54.6 & 46.0 & 54.3 & 53.0 & 51.1 \\
EuroLLM-9B \textit{(new)} & 60.4 & 61.1 & 62.2 & 61.9 & 58.1 & 57.5 & 61.8 & 61.4 & 62.0 & 60.9 & 59.1 & 60.8 & 54.2 & 60.7 & 51.4 & 59.4 & 58.2 & 57.6 \\
EuroLLM-22B \textit{(old)} & 61.4 & 61.3 & 62.4 & 62.7 & 58.5 & 58.7 & 62.2 & 62.1 & 62.4 & 62.2 & 59.5 & 61.0 & 53.4 & 61.3 & 51.3 & 59.4 & 57.9 & 56.4 \\
EuroLLM-22B \textit{(new)} & \underline{65.4} & \underline{66.4} & \underline{68.0} & \underline{67.6} & \underline{62.7} & \textbf{\underline{62.2}} & \underline{66.9} & \underline{66.4} & \underline{65.8} & \underline{66.0} & \textbf{\underline{63.6}} & \underline{66.2} & \underline{57.9} & \underline{66.2} & \underline{54.8} & \underline{64.0} & \textbf{\underline{62.5}} & \underline{61.3} \\
\bottomrule
\end{tabular}}
\caption{Per-language performance on MMMLU. \textbf{Bold} indicates the best score per benchmark within each section (\textit{Fully-open} or \textit{Open-weights}). \underline{Underlined} indicates the best \textit{Fully-open} European score.}
\label{tab:results_mmmlu}
\end{table}

\begin{table}[!htbp]
\setlength{\tabcolsep}{3pt}
\small
\resizebox{\textwidth}{!}{
\begin{tabular}{lcccccccccccccc}
\toprule
\multicolumn{1}{c}{} & \multicolumn{7}{c}{\textbf{EU}} & \multicolumn{7}{c}{\textbf{Non-EU}} \\
\cmidrule(lr){2-8} \cmidrule(lr){9-15}
\textbf{Model} & \textbf{cs} & \textbf{de} & \textbf{es} & \textbf{fr} & \textbf{hu} & \textbf{it} & \textbf{pt} & \textbf{ar} & \textbf{hi} & \textbf{ja} & \textbf{ko} & \textbf{ru} & \textbf{uk} & \textbf{zh} \\
\midrule
\multicolumn{15}{c}{\textbf{\textit{Fully-open}}} \\
\midrule
\multicolumn{15}{l}{\textbf{\textit{European}}} \\
\addlinespace[2pt]
EuroLLM-9B \textit{(old)} & 30.8 & 29.7 & 31.3 & 29.6 & 29.1 & 29.7 & 30.5 & 26.7 & 26.6 & 28.0 & 27.0 & 29.7 & 29.6 & 28.0 \\
EuroLLM-9B \textit{(new)} & 34.9 & 34.5 & 35.5 & 36.1 & 34.3 & 35.4 & 36.0 & 32.5 & 31.7 & 32.2 & 31.8 & 34.6 & 34.3 & 32.3 \\
EuroLLM-22B \textit{(old)} & 39.2 & 38.8 & 39.7 & 40.1 & 37.8 & 40.2 & 39.6 & 36.3 & 36.2 & 34.9 & 34.6 & 38.8 & 37.6 & 36.2 \\
EuroLLM-22B \textit{(new)} & \underline{46.8} & \underline{46.1} & \underline{47.5} & \underline{47.6} & \underline{45.4} & \underline{47.2} & \underline{47.2} & \underline{43.4} & \underline{42.8} & \underline{43.0} & \underline{42.8} & \underline{46.0} & \underline{46.0} & \underline{43.0} \\
\bottomrule
\end{tabular}}
\caption{Per-language performance on MMLU-ProX. \textbf{Bold} indicates the best score per benchmark within each section (\textit{Fully-open} or \textit{Open-weights}). \underline{Underlined} indicates the best \textit{Fully-open} European score.}
\label{tab:results_mmluprox}
\end{table}

\begin{table}[!htbp]
\setlength{\tabcolsep}{3pt}
\resizebox{\textwidth}{!}{
\begin{tabular}{lcccccccccccccccccc}
\toprule
\multicolumn{1}{c}{} & \multicolumn{12}{c}{\textbf{EU}} & \multicolumn{6}{c}{\textbf{Non-EU}} \\
\cmidrule(lr){2-13} \cmidrule(lr){14-19}
\textbf{Model} & \textbf{da} & \textbf{de} & \textbf{es} & \textbf{fr} & \textbf{hr} & \textbf{hu} & \textbf{it} & \textbf{nl} & \textbf{pt} & \textbf{ro} & \textbf{sk} & \textbf{sv} & \textbf{ar} & \textbf{ca} & \textbf{hi} & \textbf{ru} & \textbf{uk} & \textbf{zh} \\
\midrule
\multicolumn{19}{c}{\textbf{\textit{Fully-open}}} \\
\midrule
\multicolumn{19}{l}{\textbf{\textit{European}}} \\
\addlinespace[2pt]
EuroLLM-9B \textit{(old)} & 74.0 & 74.8 & 75.0 & 74.7 & 70.1 & 69.5 & 75.5 & 73.9 & 75.3 & 74.3 & 71.4 & 73.0 & 69.4 & 72.7 & 60.3 & 72.5 & 71.4 & 71.0 \\
EuroLLM-9B \textit{(new)}& 80.9 & 81.0 & 83.1 & 80.8 & 76.4 & 77.1 & 82.7 & 82.3 & 82.7 & 81.0 & 78.5 & 81.2 & 75.4 & 80.9 & 69.7 & 80.1 & 77.8 & 78.8 \\
EuroLLM-22B \textit{(old)} & 80.6 & 81.3 & 82.1 & 80.0 & 77.1 & 75.3 & 81.9 & 81.1 & 80.9 & 81.4 & 77.8 & 80.7 & 74.8 & 80.3 & 66.3 & 79.9 & 78.0 & 77.4 \\
EuroLLM-22B \textit{(new)} & \textbf{\underline{86.1}} & \underline{84.6} & \underline{86.2} & \underline{83.8} & \textbf{\underline{82.1}} & \textbf{\underline{80.6}} & \underline{85.8} & \textbf{\underline{84.3}} & \underline{83.8} & \textbf{\underline{85.2}} & \textbf{\underline{82.0}} & \textbf{\underline{84.5}} & \textbf{\underline{78.3}} & \textbf{\underline{84.9}} & \textbf{\underline{71.3}} & \textbf{\underline{83.0}} & \textbf{\underline{81.4}} & 80.1 \\
\bottomrule
\end{tabular}}
\caption{Per-language performance on multilingual ARC-C. \textbf{Bold} indicates the best score per benchmark within each section (\textit{Fully-open} or \textit{Open-weights}). \underline{Underlined} indicates the best \textit{Fully-open} European score.}
\label{tab:results_marcc}
\end{table}

\begin{table}[!htbp]
\setlength{\tabcolsep}{8pt}
\centering
\begin{tabular}{lcccccc}
\toprule
\multicolumn{1}{c}{} & \multicolumn{3}{c}{\textbf{EU}} & \multicolumn{3}{c}{\textbf{Non-EU}} \\
\cmidrule(lr){2-4} \cmidrule(lr){5-7}
\textbf{Model} & \textbf{de} & \textbf{es} & \textbf{fr} & \textbf{ja} & \textbf{ru} & \textbf{zh} \\
\midrule
\multicolumn{7}{c}{\textbf{\textit{Fully-open}}} \\
\midrule
\multicolumn{7}{l}{\textbf{\textit{European}}} \\
\addlinespace[2pt]
EuroLLM-9B \textit{(old)} & 61.2 & 64.4 & 60.0 & 59.6 & 63.5 & 54.5 \\
EuroLLM-9B \textit{(new)} & 67.9 & 71.2 & 70.9 & 55.6 & 71.5 & 61.7 \\
EuroLLM-22B \textit{(old)} & 72.7 & 75.2 & 73.7 & 64.0 & 75.6 & 70.0 \\
EuroLLM-22B \textit{(new)} & \underline{76.9} & \underline{77.3} & \underline{79.1} & 67.9 & \underline{81.9} & \underline{73.3} \\
Apertus-8B & 59.7 & 62.4 & 62.0 & 49.5 & 66.5 & 53.2 \\
Apertus-70B & 74.4 & 75.2 & 71.2 & \underline{68.0} & 76.8 & 70.8 \\
\addlinespace[4pt]
\multicolumn{7}{l}{\textbf{\textit{Non-European}}} \\
\addlinespace[2pt]
OLMo-3-7B & 75.6 & 83.2 & 82.9 & 61.5 & 78.7 & 78.0 \\
OLMo-3.1-32B & \textbf{88.0} & \textbf{91.1} & \textbf{87.3} & \textbf{81.3} & \textbf{93.7} & \textbf{82.8} \\
\midrule
\multicolumn{7}{c}{\textbf{\textit{Open-weights}}} \\
\midrule
\multicolumn{7}{l}{\textbf{\textit{European}}} \\
\addlinespace[2pt]
Mistral-3.2-24B & 90.7 & 92.3 & 89.3 & 83.9 & 92.5 & 88.8 \\
\addlinespace[4pt]
\multicolumn{7}{l}{\textbf{\textit{Non-European}}} \\
\addlinespace[2pt]
Llama-3.1-8B & 74.5 & 77.9 & 74.4 & 60.7 & 77.6 & 72.9 \\
Llama-3.3-70B & \textbf{92.8} & \textbf{94.0} & \textbf{92.1} & \textbf{88.8} & 92.9 & 89.2 \\
Gemma-3-12B & 88.4 & 90.7 & 83.5 & 81.7 & 87.3 & 84.7 \\
Gemma-3-27B & 88.9 & 91.6 & 89.2 & 83.6 & 90.7 & 86.7 \\
Qwen-3-14B & 90.3 & 91.2 & 89.5 & 86.9 & 92.8 & 89.6 \\
Qwen-3-32B & 92.4 & 93.1 & 90.4 & 88.5 & \textbf{94.5} & \textbf{91.2} \\
Qwen-3-30B-A3B & 91.2 & \textbf{94.0} & 88.9 & 86.1 & 92.9 & 90.0 \\
\bottomrule
\end{tabular}
\caption{Per-language performance on MGSM. \textbf{Bold} indicates the best score per benchmark within each section (\textit{Fully-open} or \textit{Open-weights}). \underline{Underlined} indicates the best \textit{Fully-open} European score.}
\label{tab:results_mgsm}
\end{table}

\begin{table}[!htbp]
\setlength{\tabcolsep}{3pt}
\resizebox{\textwidth}{!}{
\begin{tabular}{lcccccccccccccccccccccc}
\toprule
\multicolumn{1}{c}{} & \multicolumn{22}{c}{\textbf{EU}} \\
\cmidrule(lr){2-23}
\textbf{Model} & \textbf{bg} & \textbf{cs} & \textbf{da} & \textbf{de} & \textbf{el} & \textbf{es} & \textbf{et} & \textbf{fi} & \textbf{fr} & \textbf{ga} & \textbf{hr} & \textbf{hu} & \textbf{it} & \textbf{lt} & \textbf{mt} & \textbf{nl} & \textbf{pl} & \textbf{pt} & \textbf{ro} & \textbf{sk} & \textbf{sl} & \textbf{sv} \\
\midrule
\multicolumn{23}{c}{\textbf{\textit{Fully-open}}} \\
\midrule
\multicolumn{23}{l}{\textbf{\textit{European}}} \\
\addlinespace[2pt]
EuroLLM-9B \textit{(old)} & 91.7 & 92.2 & 91.7 & 88.9 & 90.0 & \textbf{\underline{87.4}} & 91.9 & 92.9 & 89.2 & 81.2 & 90.5 & \textbf{\underline{90.3}} & 89.4 & 91.1 & 72.2 & 88.7 & 90.4 & 90.4 & 91.5 & 91.4 & 90.5 & 91.6 \\
EuroLLM-9B \textit{(new)} & 91.7 & 92.0 & 91.7 & 88.7 & 90.0 & 87.2 & 92.0 & 92.7 & 88.9 & 81.1 & 90.5 & 90.1 & 89.4 & 90.9 & 71.6 & 88.7 & 90.4 & 90.1 & 91.6 & 91.3 & 90.5 & 91.6 \\
EuroLLM-22B \textit{(old)} & 91.6 & 92.2 & \textbf{\underline{91.9}} & \textbf{\underline{89.0}} & \textbf{\underline{90.3}} & \textbf{\underline{87.4}} & \textbf{\underline{92.2}} & 92.9 & 89.2 & \textbf{\underline{81.4}} & 90.8 & 90.2 & 89.5 & \textbf{\underline{91.3}} & \textbf{\underline{72.6}} & \textbf{\underline{88.9}} & \textbf{\underline{90.6}} & \textbf{\underline{90.5}} & \textbf{\underline{91.7}} & \textbf{\underline{91.6}} & \textbf{\underline{90.7}} & 91.6 \\
EuroLLM-22B \textit{(new)} & \textbf{\underline{91.8}} & \textbf{\underline{92.3}} & 91.8 & 88.9 & 90.1 & \textbf{\underline{87.4}} & 92.1 & \textbf{\underline{93.1}} & \textbf{\underline{89.3}} & \textbf{\underline{81.4}} & \textbf{\underline{91.0}} & \textbf{\underline{90.3}} & \textbf{\underline{89.6}} & 91.2 & \textbf{\underline{72.6}} & 88.8 & 90.5 & 90.3 & 91.6 & \textbf{\underline{91.6}} & \textbf{\underline{90.7}} & \textbf{\underline{91.8}} \\
\bottomrule
\end{tabular}}
\caption{FLORES performance for EU, out-of-English language pairs (en-xx). \textbf{Bold} indicates the best score per benchmark within each section (\textit{Fully-open} or \textit{Open-weights}). \underline{Underlined} indicates the best \textit{Fully-open} European score.}
\label{tab:results_flores_eu_enxx}
\end{table}

\begin{table}[!htbp]
\setlength{\tabcolsep}{3pt}
\resizebox{\textwidth}{!}{
\begin{tabular}{lcccccccccccccccccccccc}
\toprule
\multicolumn{1}{c}{} & \multicolumn{22}{c}{\textbf{EU}} \\
\cmidrule(lr){2-23}
\textbf{Model} & \textbf{bg} & \textbf{cs} & \textbf{da} & \textbf{de} & \textbf{el} & \textbf{es} & \textbf{et} & \textbf{fi} & \textbf{fr} & \textbf{ga} & \textbf{hr} & \textbf{hu} & \textbf{it} & \textbf{lt} & \textbf{mt} & \textbf{nl} & \textbf{pl} & \textbf{pt} & \textbf{ro} & \textbf{sk} & \textbf{sl} & \textbf{sv} \\
\midrule
\multicolumn{23}{c}{\textbf{\textit{Fully-open}}} \\
\midrule
\multicolumn{23}{l}{\textbf{\textit{European}}} \\
\addlinespace[2pt]
EuroLLM-9B \textit{(old)} & 88.4 & 88.7 & 90.3 & 89.4 & 88.2 & 87.2 & 89.5 & 90.1 & \textbf{\underline{89.6}} & 86.4 & 88.2 & 88.6 & 88.3 & 87.0 & 84.0 & 87.6 & 86.5 & 89.7 & 89.5 & 88.3 & 88.0 & 90.2 \\
EuroLLM-9B \textit{(new)}& \textbf{\underline{88.7}} & \textbf{\underline{89.0}} & \textbf{\underline{90.5}} & \textbf{\underline{89.6}} & \textbf{\underline{88.3}} & \textbf{\underline{87.4}} & \textbf{\underline{89.6}} & \textbf{\underline{90.3}} & \textbf{\underline{89.6}} & \textbf{\underline{86.7}} & \textbf{\underline{88.5}} & \textbf{\underline{88.8}} & \textbf{\underline{88.4}} & \textbf{\underline{87.4}} & 83.9 & \textbf{\underline{87.7}} & \textbf{\underline{86.6}} & \textbf{\underline{89.9}} & \textbf{\underline{89.8}} & \textbf{\underline{88.6}} & \textbf{\underline{88.1}} & \textbf{\underline{90.3}} \\
EuroLLM-22B \textit{(old)} & 88.5 & 88.6 & \textbf{\underline{90.5}} & \textbf{\underline{89.6}} & 88.2 & 87.2 & 89.5 & \textbf{\underline{90.3}} & 89.5 & \textbf{\underline{86.7}} & 88.3 & 88.6 & 88.1 & 87.2 & \textbf{\underline{84.2}} & 87.5 & 86.4 & 89.7 & 89.4 & 88.4 & 88.0 & 90.2 \\
EuroLLM-22B \textit{(new)}& 88.4 & 88.6 & 90.3 & 89.3 & 87.8 & 87.2 & 89.4 & 90.1 & 89.5 & 86.4 & 88.3 & 88.5 & 88.2 & 87.0 & 84.0 & 87.5 & 86.4 & 89.8 & 89.5 & 88.3 & 88.0 & \textbf{\underline{90.3}} \\
\bottomrule
\end{tabular}}
\caption{FLORES performance for EU, into-English language pairs (xx-en). \textbf{Bold} indicates the best score per benchmark within each section (\textit{Fully-open} or \textit{Open-weights}). \underline{Underlined} indicates the best \textit{Fully-open} European score.}
\label{tab:results_flores_eu_xxen}
\end{table}

\begin{table}[!htbp]
\setlength{\tabcolsep}{3pt}
\resizebox{\textwidth}{!}{
\begin{tabular}{lcccccccccccccccccc}
\toprule
\multicolumn{1}{c}{} & \multicolumn{9}{c}{\textbf{en-xx}} & \multicolumn{9}{c}{\textbf{xx-en}} \\
\cmidrule(lr){2-10} \cmidrule(lr){11-19}
\textbf{Model} & \textbf{ca} & \textbf{gl} & \textbf{hi} & \textbf{ja} & \textbf{ko} & \textbf{ru} & \textbf{tr} & \textbf{uk} & \textbf{zh} & \textbf{ca} & \textbf{gl} & \textbf{hi} & \textbf{ja} & \textbf{ko} & \textbf{ru} & \textbf{tr} & \textbf{uk} & \textbf{zh} \\
\midrule
\multicolumn{19}{c}{\textbf{\textit{Fully-open}}} \\
\midrule
\multicolumn{19}{l}{\textbf{\textit{European}}} \\
\addlinespace[2pt]
EuroLLM-9B \textit{(old)} & 88.3 & 88.4 & 80.7 & 91.7 & 89.9 & \textbf{\underline{90.4}} & 90.2 & 90.6 & 88.8 & \textbf{\underline{89.2}} & \textbf{\underline{88.8}} & 89.4 & 88.4 & \textbf{\underline{88.6}} & 86.9 & 89.7 & 87.5 & 87.4 \\
EuroLLM-9B \textit{(new)}& 88.0 & 88.3 & 80.8 & 91.6 & 89.9 & \textbf{\underline{90.4}} & 90.3 & 90.6 & 88.9 & \textbf{\underline{89.2}} & \textbf{\underline{88.8}} & \textbf{\underline{89.7}} & 88.3 & \textbf{\underline{88.6}} & \textbf{\underline{87.1}} & \textbf{\underline{89.8}} & \textbf{\underline{87.8}} & \textbf{\underline{87.6}} \\
EuroLLM-22B \textit{(old)} & \textbf{\underline{88.5}} & 88.5 & 80.8 & 91.7 & \textbf{\underline{90.2}} & \textbf{\underline{90.4}} & 90.5 & \textbf{\underline{90.9}} & 88.9 & 89.1 & 88.7 & 89.5 & 88.4 & 88.5 & 86.9 & 89.5 & 87.6 & 87.3 \\
EuroLLM-22B \textit{(new)} & 88.4 & \textbf{\underline{88.6}} & \textbf{\underline{81.3}} & \textbf{\underline{91.8}} & \textbf{\underline{90.2}} & \textbf{\underline{90.4}} & \textbf{\underline{90.6}} & 90.7 & \textbf{\underline{89.1}} & 89.1 & \textbf{\underline{88.8}} & 89.6 & \textbf{\underline{88.5}} & \textbf{\underline{88.6}} & 86.9 & 89.5 & 87.5 & 87.3 \\
\bottomrule
\end{tabular}}
\caption{FLORES performance for non-EU language pairs. \textbf{Bold} indicates the best score per benchmark within each section (\textit{Fully-open} or \textit{Open-weights}). \underline{Underlined} indicates the best \textit{Fully-open} European score.}
\label{tab:results_flores_non_eu}
\end{table}

\begin{table}[!htbp]
\setlength{\tabcolsep}{3pt}
\resizebox{\textwidth}{!}{
\begin{tabular}{lccccccccccccccccccccc}
\toprule
\multicolumn{1}{c}{} & \multicolumn{21}{c}{\textbf{EU}} \\
\cmidrule(lr){2-22}
\textbf{Model} & \textbf{bg} & \textbf{cs} & \textbf{da} & \textbf{de} & \textbf{el} & \textbf{es} & \textbf{et} & \textbf{fi} & \textbf{fr} & \textbf{hr} & \textbf{hu} & \textbf{it} & \textbf{lt} & \textbf{lv} & \textbf{nl} & \textbf{pl} & \textbf{pt} & \textbf{ro} & \textbf{sk} & \textbf{sl} & \textbf{sv} \\
\midrule
\multicolumn{22}{c}{\textbf{\textit{Fully-open}}} \\
\midrule
\multicolumn{22}{l}{\textbf{\textit{European}}} \\
\addlinespace[2pt]
EuroLLM-9B \textit{(old)} & 85.5 & 85.0 & 85.0 & 82.1 & 86.3 & 83.4 & 86.5 & 66.1 & 81.3 & 83.8 & 83.9 & 84.5 & 84.5 & 84.4 & 84.2 & 84.2 & 84.2 & 85.7 & 83.7 & 83.9 & 86.0 \\
EuroLLM-9B \textit{(new)} & 85.5 & 84.9 & 85.1 & 82.5 & 86.1 & 83.1 & 86.6 & 66.1 & 81.5 & 84.1 & 83.9 & 84.3 & 84.8 & 84.2 & 84.3 & 84.3 & 83.8 & 85.5 & 83.9 & 83.8 & 86.2 \\
EuroLLM-22B \textit{(old)} & 86.0 & \textbf{\underline{85.8}} & \textbf{\underline{85.6}} & \textbf{\underline{82.7}} & 86.5 & \textbf{\underline{83.8}} & 86.9 & \textbf{\underline{66.4}} & 81.4 & 84.6 & 84.0 & 84.6 & \textbf{\underline{85.1}} & \textbf{\underline{85.0}} & \textbf{\underline{84.4}} & 84.7 & \textbf{\underline{84.6}} & 85.9 & \textbf{\underline{84.1}} & \textbf{\underline{84.4}} & 86.4 \\
EuroLLM-22B \textit{(new)} & \textbf{\underline{86.1}} & 85.7 & 85.4 & 82.5 & \textbf{\underline{86.6}} & 83.7 & \textbf{\underline{87.1}} & 66.3 & \textbf{\underline{81.6}} & \textbf{\underline{85.1}} & \textbf{\underline{84.2}} & \textbf{\underline{84.7}} & 84.9 & 84.6 & \textbf{\underline{84.4}} & \textbf{\underline{84.9}} & 83.9 & \textbf{\underline{86.0}} & \textbf{\underline{84.1}} & \textbf{\underline{84.4}} & \textbf{\underline{86.6}} \\
\bottomrule
\end{tabular}}
\caption{WMT24++ performance for EU, out-of-English language pairs (en-xx). \textbf{Bold} indicates the best score per benchmark within each section (\textit{Fully-open} or \textit{Open-weights}). \underline{Underlined} indicates the best \textit{Fully-open} European score.}
\label{tab:results_wmt24pp_eu_enxx}
\end{table}

\begin{table}[!htbp]
\setlength{\tabcolsep}{3pt}
\resizebox{\textwidth}{!}{
\begin{tabular}{lccccccccccccccccccccc}
\toprule
\multicolumn{1}{c}{} & \multicolumn{21}{c}{\textbf{EU}} \\
\cmidrule(lr){2-22}
\textbf{Model} & \textbf{bg} & \textbf{cs} & \textbf{da} & \textbf{de} & \textbf{el} & \textbf{es} & \textbf{et} & \textbf{fi} & \textbf{fr} & \textbf{hr} & \textbf{hu} & \textbf{it} & \textbf{lt} & \textbf{lv} & \textbf{nl} & \textbf{pl} & \textbf{pt} & \textbf{ro} & \textbf{sk} & \textbf{sl} & \textbf{sv} \\
\midrule
\multicolumn{22}{c}{\textbf{\textit{Fully-open}}} \\
\midrule
\multicolumn{22}{l}{\textbf{\textit{European}}} \\
\addlinespace[2pt]
EuroLLM-9B \textit{(old)} & 84.0 & 83.7 & 85.5 & 84.3 & 85.2 & 84.9 & 85.7 & 70.7 & \textbf{\underline{84.1}} & 83.5 & 83.4 & 84.9 & 81.9 & 83.1 & 84.9 & 82.0 & 84.5 & 84.7 & 83.0 & 83.6 & 86.1 \\
EuroLLM-9B \textit{(new)} & 84.1 & \textbf{\underline{83.8}} & 85.7 & 84.5 & 85.4 & 84.9 & 85.7 & 71.1 & \textbf{\underline{84.1}} & 83.9 & \textbf{\underline{83.6}} & \textbf{\underline{85.1}} & 82.0 & 83.1 & 84.9 & \textbf{\underline{82.3}} & 84.7 & \textbf{\underline{84.9}} & \textbf{\underline{83.4}} & 83.5 & 86.3 \\
EuroLLM-22B \textit{(old)} & \textbf{\underline{84.2}} & 83.7 & \textbf{\underline{85.8}} & \textbf{\underline{84.6}} & \textbf{\underline{85.6}} & \textbf{\underline{85.2}} & \textbf{\underline{86.2}} & 73.9 & \textbf{\underline{84.1}} & \textbf{\underline{84.1}} & \textbf{\underline{83.6}} & 85.0 & \textbf{\underline{82.2}} & \textbf{\underline{83.4}} & \textbf{\underline{85.1}} & 82.2 & \textbf{\underline{85.0}} & 84.8 & 83.3 & \textbf{\underline{83.9}} & \textbf{\underline{86.4}} \\
EuroLLM-22B \textit{(new)} & 83.9 & \textbf{\underline{83.8}} & 85.5 & 84.3 & 85.1 & 85.1 & 85.7 & 74.6 & \textbf{\underline{84.1}} & 84.0 & 83.4 & 84.9 & \textbf{\underline{82.2}} & 83.1 & 85.0 & 82.2 & 84.8 & \textbf{\underline{84.9}} & 83.0 & 83.6 & 86.3 \\
\bottomrule
\end{tabular}}
\caption{WMT24++ performance for EU, into-English language pairs (xx-en). \textbf{Bold} indicates the best score per benchmark within each section (\textit{Fully-open} or \textit{Open-weights}). \underline{Underlined} indicates the best \textit{Fully-open} European score.}
\label{tab:results_wmt24pp_eu_xxen}
\end{table}

\begin{table}[!htbp]
\setlength{\tabcolsep}{3pt}
\resizebox{\textwidth}{!}{
\begin{tabular}{lcccccccccccccccccccc}
\toprule
\multicolumn{1}{c}{} & \multicolumn{10}{c}{\textbf{en-xx}} & \multicolumn{10}{c}{\textbf{xx-en}} \\
\cmidrule(lr){2-11} \cmidrule(lr){12-21}
\textbf{Model} & \textbf{ar} & \textbf{ca} & \textbf{hi} & \textbf{ja} & \textbf{ko} & \textbf{no} & \textbf{ru} & \textbf{tr} & \textbf{uk} & \textbf{zh} & \textbf{ar} & \textbf{ca} & \textbf{hi} & \textbf{ja} & \textbf{ko} & \textbf{no} & \textbf{ru} & \textbf{tr} & \textbf{uk} & \textbf{zh} \\
\midrule
\multicolumn{21}{c}{\textbf{\textit{Fully-open}}} \\
\midrule
\multicolumn{21}{l}{\textbf{\textit{European}}} \\
\addlinespace[2pt]
EuroLLM-9B \textit{(old)} & 77.4 & 81.3 & 70.5 & 85.8 & 85.7 & 86.7 & 81.8 & 83.9 & 84.2 & 83.5 & 78.5 & 83.6 & 83.6 & 82.8 & 83.9 & 86.5 & 79.7 & 85.0 & 82.6 & 82.3 \\
EuroLLM-9B \textit{(new)} & 77.6 & 81.1 & \textbf{\underline{71.9}} & 85.9 & 85.7 & 86.2 & 82.0 & 84.0 & 84.7 & 80.1 & 78.7 & 83.7 & 83.7 & 83.0 & \textbf{\underline{84.1}} & \textbf{\underline{86.8}} & 79.9 & \textbf{\underline{85.1}} & \textbf{\underline{82.8}} & \textbf{\underline{82.7}} \\
EuroLLM-22B \textit{(old)} & 77.7 & \textbf{\underline{81.9}} & 71.7 & 85.8 & \textbf{\underline{86.3}} & 86.7 & \textbf{\underline{82.3}} & \textbf{\underline{84.5}} & 85.0 & \textbf{\underline{83.9}} & \textbf{\underline{79.0}} & \textbf{\underline{83.8}} & 83.7 & \textbf{\underline{83.1}} & 84.0 & \textbf{\underline{86.8}} & \textbf{\underline{80.2}} & 84.8 & \textbf{\underline{82.8}} & 82.6 \\
EuroLLM-22B \textit{(new)} & \textbf{\underline{77.9}} & \textbf{\underline{81.9}} & 71.4 & \textbf{\underline{86.5}} & 86.2 & \textbf{\underline{86.8}} & 82.0 & 84.2 & \textbf{\underline{85.2}} & \textbf{\underline{83.9}} & 78.9 & 83.6 & \textbf{\underline{84.0}} & 83.0 & \textbf{\underline{84.1}} & 86.4 & 80.1 & 84.9 & 82.5 & 82.6 \\
\bottomrule
\end{tabular}}
\caption{WMT24++ performance for non-EU language pairs. \textbf{Bold} indicates the best score per benchmark within each section (\textit{Fully-open} or \textit{Open-weights}). \underline{Underlined} indicates the best \textit{Fully-open} European score.}
\label{tab:results_wmt24pp_non_eu}
\end{table}

\begin{table}[!htbp]
\setlength{\tabcolsep}{2.5pt}
\centering
\resizebox{\textwidth}{!}{
\begin{tabular}{lccccccccccc}
\toprule
\multicolumn{1}{c}{} & \multicolumn{3}{c}{\textbf{EU}} & \multicolumn{8}{c}{\textbf{Non-EU}} \\
\cmidrule(lr){2-4} \cmidrule(lr){5-12}
\textbf{Model} & \textbf{en-cs} & \textbf{en-et} & \textbf{cs-de} & \textbf{en-ar} & \textbf{en-ja} & \textbf{en-ko} & \textbf{en-ru} & \textbf{en-uk} & \textbf{en-zh} & \textbf{cs-uk} & \textbf{ja-zh} \\
\midrule
\multicolumn{12}{c}{\textbf{\textit{Fully-open}}} \\
\midrule
\multicolumn{12}{l}{\textbf{\textit{European}}} \\
\addlinespace[2pt]
EuroLLM-9B \textit{(old)} & * & * & * & * & * & * & * & * & * & * & * \\
EuroLLM-9B \textit{(new)} & \textbf{\underline{82.0}} & 80.0 & 78.8 & 71.7 & 83.3 & 80.5 & 80.2 & 80.9 & 80.1 & 84.5 & \textbf{\underline{79.6}} \\
EuroLLM-22B \textit{(old)} & * & * & * & * & * & * & * & * & * & * & * \\
EuroLLM-22B \textit{(new)} & 81.7 & 80.1 & 78.0 & 70.1 & 82.8 & 81.4 & 78.9 & 79.5 & 79.6 & \textbf{\underline{84.9}} & 79.5 \\
\bottomrule
\end{tabular}}
\caption{WMT25 results by language pair. \textbf{Bold} indicates the best score per benchmark within each section (\textit{Fully-open} or \textit{Open-weights}). \underline{Underlined} indicates the best \textit{Fully-open} European score. \textit{*Not evaluated because the required context exceeds the model's maximum context length.}}
\label{tab:results_wmt25}
\end{table}

\clearpage

\section{Results for Base Models}
\label{sec:base_model_evaluation}

To avoid the answer-formatting issues inherent to base models, we evaluate them on multiple-choice benchmarks---HellaSwag, MMLU, ARC-C, and their multilingual counterparts---using a 3-shot likelihood-based approach. 
For each question, the candidate choices are concatenated to the question one at a time, and the log-likelihood is computed for each resulting sequence. 
The model predicts the choice with the highest log-likelihood, which we compare to the ground-truth answer.
As baselines, we use the base versions of the instruct models discussed in \autoref{sec:evaluation}, whenever available.

\subsection{Aggregated Results}

\begin{table}[!htbp]
\center\small
\small
\begin{tabular}{lccc}
\toprule
\textbf{Model} & \textbf{Hellaswag} & \textbf{MMLU} & \textbf{ARC-C} \\
\midrule
\multicolumn{4}{c}{\textbf{\textit{Fully-open}}} \\
\midrule
\multicolumn{4}{l}{\textbf{\textit{European}}} \\
\addlinespace[2pt]
EuroLLM-9B-Base & 73.6 & 43.9 & 58.8 \\
EuroLLM-22B-Base & 73.2 & 46.4 & 62.3 \\
Apertus-8B-Base & 73.2 & 47.1 & 63.1 \\
Apertus-70B-Base & \textbf{\underline{77.4}} & \underline{49.4} & \underline{64.0} \\
\addlinespace[4pt]
\multicolumn{4}{l}{\textbf{\textit{Non-European}}} \\
\addlinespace[2pt]
OLMo-3-7B-Base & 69.2 & 46.0 & 61.8 \\
OLMo-3-32B-Base & 77.2 & \textbf{52.0} & \textbf{67.9} \\
\midrule
\multicolumn{4}{c}{\textbf{\textit{Open-weights}}} \\
\midrule
\multicolumn{4}{l}{\textbf{\textit{European}}} \\
\addlinespace[2pt]
Mistral-3.2-24B-Base & 79.3 & 53.9 & 68.5 \\
\addlinespace[4pt]
\multicolumn{4}{l}{\textbf{\textit{Non-European}}} \\
\addlinespace[2pt]
Llama-3.1-8B-Base & 75.7 & 46.3 & 58.3 \\
Llama-3.3-70B-Base & \textbf{83.9} & \textbf{54.9} & 68.4 \\
Gemma-3-12B-Base & 77.7 & 51.9 & 68.2 \\
Gemma-3-27B-Base & 78.2 & 54.7 & \textbf{70.5} \\
Qwen-3-14B-Base & 76.2 & 54.2 & 68.3 \\
Qwen-3-30B-A3B-Base & 76.5 & 53.0 & 60.6 \\
\bottomrule
\end{tabular}
\caption{Results on English benchmarks. \textbf{Bold} indicates the best score per benchmark within each section (\textit{Fully-open} or \textit{Open-weights}). \underline{Underlined} indicates the best \textit{Fully-open} European score.}
\label{tab:base_results_en}
\end{table}

\begin{table}[!htbp]
\small
\centering
\begin{tabular}{lccc}
\toprule
\textbf{Model} & \textbf{Hellaswag} & \textbf{MMMLU} & \textbf{ARC-C} \\
\midrule
\multicolumn{4}{c}{\textbf{\textit{Fully-open}}} \\
\midrule
\multicolumn{4}{l}{\textbf{\textit{European}}} \\
\addlinespace[2pt]
EuroLLM-9B-Base & 61.1 & 39.1 & 50.7 \\
EuroLLM-22B-Base & 62.9 & 41.5 & 53.1 \\
Apertus-8B-Base & 63.6 & 40.9 & 52.6 \\
Apertus-70B-Base & \textbf{\underline{67.6}} & \underline{42.3} & \underline{54.5} \\
\addlinespace[4pt]
\multicolumn{4}{l}{\textbf{\textit{Non-European}}} \\
\addlinespace[2pt]
OLMo-3-7B-Base & 40.8 & 32.6 & 34.0 \\
OLMo-3-32B-Base & 54.2 & \textbf{39.3} & \textbf{46.9} \\
\midrule
\multicolumn{4}{c}{\textbf{\textit{Open-weights}}} \\
\midrule
\multicolumn{4}{l}{\textbf{\textit{European}}} \\
\addlinespace[2pt]
Mistral-3.2-24B-Base & 66.4 & 46.1 & 58.0 \\
\addlinespace[4pt]
\multicolumn{4}{l}{\textbf{\textit{Non-European}}} \\
\addlinespace[2pt]
Llama-3.1-8B-Base & 56.4 & 37.0 & 43.8 \\
Llama-3.3-70B-Base & \textbf{69.3} & 46.6 & 57.7 \\
Gemma-3-12B-Base & 65.7 & 44.6 & 57.9 \\
Gemma-3-27B-Base & 68.9 & \textbf{48.3} & \textbf{60.6} \\
Qwen-3-14B-Base & 61.2 & 41.2 & 54.3 \\
Qwen-3-30B-A3B-Base & 62.6 & 40.9 & 54.5 \\
\bottomrule
\end{tabular}
\caption{Results on multilingual benchmarks. \textbf{Bold} indicates the best score per benchmark within each section (\textit{Fully-open} or \textit{Open-weights}). \underline{Underlined} indicates the best \textit{Fully-open} European score.}
\label{tab:base_results_xx}
\end{table}

\begin{table}[!htbp]
\small
\centering
\begin{tabular}{lccc}
\toprule
\textbf{Model} & \textbf{Hellaswag} & \textbf{MMMLU} & \textbf{ARC-C} \\
\midrule
\multicolumn{4}{c}{\textbf{\textit{Fully-open}}} \\
\midrule
\multicolumn{4}{l}{\textbf{\textit{European}}} \\
\addlinespace[2pt]
EuroLLM-9B-Base & 63.0 & 40.2 & 52.4 \\
EuroLLM-22B-Base & 64.8 & 42.5 & 54.9 \\
Apertus-8B-Base & 65.5 & 41.9 & 54.4 \\
Apertus-70B-Base & \textbf{\underline{69.7}} & \underline{43.6} & \underline{56.7} \\
\addlinespace[4pt]
\multicolumn{4}{l}{\textbf{\textit{Non-European}}} \\
\addlinespace[2pt]
OLMo-3-7B-Base & 42.0 & 33.6 & 35.1 \\
OLMo-3-32B-Base & 55.9 & \textbf{40.5} & \textbf{48.6} \\
\midrule
\multicolumn{4}{c}{\textbf{\textit{Open-weights}}} \\
\midrule
\multicolumn{4}{l}{\textbf{\textit{European}}} \\
\addlinespace[2pt]
Mistral-3.2-24B-Base & 68.6 & 47.5 & 60.4 \\
\addlinespace[4pt]
\multicolumn{4}{l}{\textbf{\textit{Non-European}}} \\
\addlinespace[2pt]
Llama-3.1-8B-Base & 57.9 & 38.0 & 45.0 \\
Llama-3.3-70B-Base & \textbf{71.1} & 47.8 & 59.3 \\
Gemma-3-12B-Base & 67.6 & 45.7 & 59.7 \\
Gemma-3-27B-Base & 70.8 & \textbf{49.5} & \textbf{62.7} \\
Qwen-3-14B-Base & 62.7 & 45.1 & 55.8 \\
Qwen-3-30B-A3B-Base & 64.3 & 44.8 & 55.9 \\
\bottomrule
\end{tabular}
\caption{Average performance on European languages. \textbf{Bold} indicates the best score per benchmark within each section (\textit{Fully-open} or \textit{Open-weights}). \underline{Underlined} indicates the best \textit{Fully-open} European score.}
\label{tab:base_results_eu}
\end{table}

\begin{table}[!htbp]
\small
\centering
\begin{tabular}{lccc}
\toprule
\textbf{Model} & \textbf{Hellaswag} & \textbf{MMMLU} & \textbf{ARC-C} \\
\midrule
\multicolumn{4}{c}{\textbf{\textit{Fully-open}}} \\
\midrule
\multicolumn{4}{l}{\textbf{\textit{European}}} \\
\addlinespace[2pt]
EuroLLM-9B-Base & 56.6 & 37.1 & 47.4 \\
EuroLLM-22B-Base & 58.5 & 39.4 & 49.5 \\
Apertus-8B-Base & 58.9 & 38.7 & 48.9 \\
Apertus-70B-Base & \textbf{\underline{62.7}} & \textbf{\underline{39.7}} & \underline{49.9} \\
\addlinespace[4pt]
\multicolumn{4}{l}{\textbf{\textit{Non-European}}} \\
\addlinespace[2pt]
OLMo-3-7B-Base & 38.0 & 30.8 & 31.9 \\
OLMo-3-32B-Base & 50.1 & 36.7 & \textbf{43.5} \\
\midrule
\multicolumn{4}{c}{\textbf{\textit{Open-weights}}} \\
\midrule
\multicolumn{4}{l}{\textbf{\textit{European}}} \\
\addlinespace[2pt]
Mistral-3.2-24B-Base & 60.9 & 43.4 & 53.2 \\
\addlinespace[4pt]
\multicolumn{4}{l}{\textbf{\textit{Non-European}}} \\
\addlinespace[2pt]
Llama-3.1-8B-Base & 52.7 & 35.0 & 41.4 \\
Llama-3.3-70B-Base & \textbf{65.2} & 44.3 & 54.6 \\
Gemma-3-12B-Base & 61.2 & 42.5 & 54.3 \\
Gemma-3-27B-Base & 64.3 & \textbf{46.0} & \textbf{56.6} \\
Qwen-3-14B-Base & 57.5 & 33.5 & 51.3 \\
Qwen-3-30B-A3B-Base & 58.6 & 33.0 & 51.6 \\
\bottomrule
\end{tabular}
\caption{Average performance on non-European languages. \textbf{Bold} indicates the best score per benchmark within each section (\textit{Fully-open} or \textit{Open-weights}). \underline{Underlined} indicates the best \textit{Fully-open} European score.}
\label{tab:base_results_non_eu}
\end{table}

\clearpage
\subsection{Per-Language Multilingual Results}

\vspace*{\fill}

\begin{table}[H]
\setlength{\tabcolsep}{3pt}
\resizebox{\textwidth}{!}{
\begin{tabular}{lccccccccccccccccc}
\toprule
\multicolumn{1}{c}{} & \multicolumn{12}{c}{\textbf{EU}} & \multicolumn{5}{c}{\textbf{Non-EU}} \\
\cmidrule(lr){2-13} \cmidrule(lr){14-18}
\textbf{Model} & \textbf{da} & \textbf{de} & \textbf{es} & \textbf{fr} & \textbf{hr} & \textbf{hu} & \textbf{it} & \textbf{nl} & \textbf{pt} & \textbf{ro} & \textbf{sk} & \textbf{sv} & \textbf{ar} & \textbf{ca} & \textbf{hi} & \textbf{ru} & \textbf{uk} \\
\midrule
\multicolumn{18}{c}{\textbf{\textit{Fully-open}}} \\
\midrule
\multicolumn{18}{l}{\textbf{\textit{European}}} \\
\addlinespace[2pt]
EuroLLM-9B-Base & 65.1 & 63.6 & 66.5 & 65.9 & 59.1 & 54.2 & 65.2 & 65.4 & 64.9 & 61.4 & 59.0 & 65.2 & 55.8 & 60.7 & 48.5 & 59.6 & 58.2 \\
EuroLLM-22B-Base & 68.5 & 64.5 & 67.6 & 68.1 & 62.5 & 55.8 & 66.2 & 66.8 & 65.6 & 63.6 & 61.8 & 66.5 & 57.1 & 62.9 & 50.4 & 61.2 & 60.9 \\
Apertus-8B-Base & 68.2 & 65.9 & 69.1 & 68.5 & 63.4 & 56.1 & 67.4 & 67.6 & 67.5 & 63.8 & 61.5 & 67.5 & 56.0 & 63.7 & 50.0 & 63.4 & 61.3 \\
Apertus-70B-Base & \textbf{\underline{72.5}} & \textbf{\underline{70.7}} & \textbf{\underline{73.0}} & \textbf{\underline{72.9}} & \textbf{\underline{67.3}} & \textbf{\underline{60.2}} & \textbf{\underline{71.5}} & \textbf{\underline{71.9}} & \textbf{\underline{71.9}} & \textbf{\underline{67.4}} & \textbf{\underline{64.8}} & \textbf{\underline{71.9}} & \textbf{\underline{60.1}} & \textbf{\underline{68.3}} & \textbf{\underline{51.6}} & \textbf{\underline{68.1}} & \textbf{\underline{65.4}} \\
\addlinespace[4pt]
\multicolumn{18}{l}{\textbf{\textit{Non-European}}} \\
\addlinespace[2pt]
OLMo-3-7B-Base & 39.4 & 45.1 & 51.8 & 51.6 & 34.7 & 31.6 & 45.5 & 41.7 & 49.6 & 39.2 & 33.5 & 40.6 & 35.4 & 41.2 & 30.7 & 45.1 & 37.5 \\
OLMo-3-32B-Base & 55.7 & 59.8 & 64.9 & 65.0 & 48.3 & 39.3 & 61.3 & 58.3 & 63.7 & 53.4 & 44.5 & 56.1 & 47.8 & 54.4 & 39.4 & 58.6 & 50.1 \\
\midrule
\multicolumn{18}{c}{\textbf{\textit{Open-weights}}} \\
\midrule
\multicolumn{18}{l}{\textbf{\textit{European}}} \\
\addlinespace[2pt]
Mistral-3.2-24B-Base & 70.1 & 71.2 & 73.9 & \textbf{73.9} & 64.6 & 55.2 & 72.2 & 70.3 & 73.1 & 66.0 & 62.1 & 71.2 & 59.2 & 68.0 & 45.1 & 67.3 & 65.0 \\
\addlinespace[4pt]
\multicolumn{18}{l}{\textbf{\textit{Non-European}}} \\
\addlinespace[2pt]
Llama-3.1-8B-Base & 57.7 & 59.3 & 64.3 & 63.4 & 51.9 & 48.7 & 61.2 & 60.9 & 63.4 & 54.6 & 49.9 & 59.6 & 48.9 & 58.8 & 45.4 & 56.5 & 54.0 \\
Llama-3.3-70B-Base & \textbf{73.6} & \textbf{72.1} & \textbf{74.9} & 73.8 & 68.1 & 61.6 & \textbf{73.1} & \textbf{74.5} & \textbf{73.8} & 67.7 & 65.4 & \textbf{74.1} & \textbf{63.1} & \textbf{71.1} & \textbf{57.4} & \textbf{67.9} & 66.7 \\
Gemma-3-12B-Base & 71.3 & 67.0 & 70.4 & 70.2 & 66.1 & 58.2 & 69.3 & 70.0 & 69.1 & 65.8 & 63.4 & 70.5 & 60.4 & 65.4 & 52.4 & 64.2 & 63.7 \\
Gemma-3-27B-Base & \textbf{73.6} & 70.7 & 73.7 & 73.8 & \textbf{69.4} & \textbf{61.7} & 72.1 & 73.7 & 71.8 & \textbf{69.1} & \textbf{66.2} & 73.5 & 62.9 & 69.3 & 55.1 & 67.3 & \textbf{67.0} \\
Qwen-3-14B-Base & 61.9 & 65.0 & 68.3 & 68.2 & 57.5 & 52.9 & 67.2 & 64.3 & 68.5 & 60.5 & 55.8 & 62.4 & 56.3 & 61.9 & 48.3 & 62.2 & 58.9 \\
Qwen-3-30B-A3B-Base & 63.8 & 65.8 & 69.9 & 69.4 & 60.0 & 54.5 & 68.6 & 65.5 & 69.1 & 61.9 & 58.1 & 65.0 & 57.6 & 64.0 & 48.1 & 62.9 & 60.5 \\
\bottomrule
\end{tabular}}
\caption{Per-language performance on multilingual Hellaswag. \textbf{Bold} indicates the best score per benchmark within each section (\textit{Fully-open} or \textit{Open-weights}). \underline{Underlined} indicates the best \textit{Fully-open} European score.}
\label{tab:base_results_mhellaswag}
\end{table}

\vspace*{\fill}

\begin{table}[H]
\setlength{\tabcolsep}{3pt}
\resizebox{\textwidth}{!}{
\begin{tabular}{lcccccccccccccccccc}
\toprule
\multicolumn{1}{c}{} & \multicolumn{12}{c}{\textbf{EU}} & \multicolumn{6}{c}{\textbf{Non-EU}} \\
\cmidrule(lr){2-13} \cmidrule(lr){14-19}
\textbf{Model} & \textbf{da} & \textbf{de} & \textbf{es} & \textbf{fr} & \textbf{hr} & \textbf{hu} & \textbf{it} & \textbf{nl} & \textbf{pt} & \textbf{ro} & \textbf{sk} & \textbf{sv} & \textbf{ar} & \textbf{ca} & \textbf{hi} & \textbf{ru} & \textbf{uk} & \textbf{zh} \\
\midrule
\multicolumn{19}{c}{\textbf{\textit{Fully-open}}} \\
\midrule
\multicolumn{19}{l}{\textbf{\textit{European}}} \\
\addlinespace[2pt]
EuroLLM-9B-Base & 41.0 & 40.5 & 41.6 & 42.3 & 38.4 & 37.3 & 41.3 & 40.5 & 41.1 & 39.5 & 38.5 & 40.1 & 35.1 & 39.0 & 33.4 & 38.7 & 37.7 & 38.6 \\
EuroLLM-22B-Base & 43.5 & 43.0 & 44.0 & 44.4 & 40.6 & 39.3 & 43.6 & 42.9 & 43.2 & 42.0 & \textbf{\underline{41.4}} & 42.7 & \textbf{\underline{37.3}} & 42.1 & \textbf{\underline{35.1}} & 41.2 & 40.0 & \textbf{\underline{40.7}} \\
Apertus-8B-Base & 42.5 & 43.6 & 43.7 & 43.5 & 40.6 & 38.1 & 43.0 & 42.2 & 43.5 & 41.0 & 39.7 & 41.6 & 36.8 & 41.5 & 34.1 & 40.4 & 39.6 & 40.0 \\
Apertus-70B-Base & \textbf{\underline{44.9}} & \textbf{\underline{45.1}} & \textbf{\underline{45.6}} & \textbf{\underline{45.4}} & \textbf{\underline{42.0}} & \textbf{\underline{39.9}} & \textbf{\underline{44.8}} & \textbf{\underline{43.5}} & \textbf{\underline{44.8}} & \textbf{\underline{42.7}} & 41.1 & \textbf{\underline{43.4}} & 37.2 & \textbf{\underline{43.7}} & 33.8 & \textbf{\underline{42.4}} & \textbf{\underline{40.4}} & 40.6 \\
\addlinespace[4pt]
\multicolumn{19}{l}{\textbf{\textit{Non-European}}} \\
\addlinespace[2pt]
OLMo-3-7B-Base & 33.2 & 35.4 & 36.0 & 36.8 & 30.4 & 29.6 & 34.9 & 34.6 & 35.5 & 32.2 & 31.4 & 32.9 & 28.7 & 33.7 & 28.2 & 31.9 & 30.2 & 32.1 \\
OLMo-3-32B-Base & 40.5 & 42.8 & 43.7 & 43.7 & 37.4 & 34.5 & 42.8 & 41.6 & 43.1 & 38.7 & 37.2 & 40.4 & 33.3 & 40.7 & 31.9 & 38.9 & 36.4 & 39.1 \\
\midrule
\multicolumn{19}{c}{\textbf{\textit{Open-weights}}} \\
\midrule
\multicolumn{19}{l}{\textbf{\textit{European}}} \\
\addlinespace[2pt]
Mistral-3.2-24B-Base & 47.8 & 49.3 & 49.2 & 50.2 & 45.4 & 42.6 & \textbf{50.6} & 47.3 & 50.1 & 46.1 & 43.6 & 48.1 & 41.2 & 47.6 & 35.1 & 47.4 & 44.2 & 44.8 \\
\addlinespace[4pt]
\multicolumn{19}{l}{\textbf{\textit{Non-European}}} \\
\addlinespace[2pt]
Llama-3.1-8B-Base & 38.3 & 39.2 & 39.9 & 39.8 & 36.0 & 35.5 & 39.7 & 38.9 & 39.4 & 36.3 & 35.0 & 37.6 & 32.2 & 38.9 & 31.5 & 36.7 & 35.1 & 35.6 \\
Llama-3.3-70B-Base & 47.7 & 48.5 & 49.4 & 50.9 & 45.5 & 44.7 & 50.3 & 48.6 & 49.8 & 46.0 & 44.6 & 47.3 & 41.9 & 48.6 & 40.0 & 45.9 & 44.3 & 44.8 \\
Gemma-3-12B-Base & 46.5 & 46.4 & 46.9 & 47.1 & 44.8 & 42.2 & 47.1 & 46.5 & 46.9 & 44.7 & 43.9 & 45.5 & 39.8 & 45.7 & 38.2 & 44.2 & 43.3 & 43.6 \\
Gemma-3-27B-Base & \textbf{50.6} & \textbf{49.7} & \textbf{50.5} & \textbf{51.4} & \textbf{48.7} & \textbf{46.4} & \textbf{50.6} & \textbf{49.6} & \textbf{50.8} & \textbf{48.2} & \textbf{47.7} & \textbf{49.3} & \textbf{43.9} & \textbf{48.9} & \textbf{41.0} & \textbf{47.9} & \textbf{46.9} & \textbf{47.2} \\
Qwen-3-14B-Base & 44.8 & 46.3 & 47.1 & 48.5 & 42.8 & 41.1 & 47.2 & 45.0 & 48.0 & 43.3 & 43.1 & 44.2 & 39.4 & 46.0 & 24.8 & 22.7 & 22.7 & 45.4 \\
Qwen-3-30B-A3B-Base & 45.3 & 46.7 & 46.7 & 47.3 & 40.6 & 40.8 & 47.6 & 45.8 & 46.8 & 42.8 & 43.1 & 44.5 & 39.0 & 44.9 & 24.7 & 22.7 & 22.7 & 44.0 \\
\bottomrule
\end{tabular}}
\caption{Per-language performance on MMMLU. \textbf{Bold} indicates the best score per benchmark within each section (\textit{Fully-open} or \textit{Open-weights}). \underline{Underlined} indicates the best \textit{Fully-open} European score.}
\label{tab:base_results_mmmlu}
\end{table}

\vspace*{\fill}

\begin{table}[!htbp]
\setlength{\tabcolsep}{3pt}
\resizebox{\textwidth}{!}{
\begin{tabular}{lcccccccccccccccccc}
\toprule
\multicolumn{1}{c}{} & \multicolumn{12}{c}{\textbf{EU}} & \multicolumn{6}{c}{\textbf{Non-EU}} \\
\cmidrule(lr){2-13} \cmidrule(lr){14-19}
\textbf{Model} & \textbf{da} & \textbf{de} & \textbf{es} & \textbf{fr} & \textbf{hr} & \textbf{hu} & \textbf{it} & \textbf{nl} & \textbf{pt} & \textbf{ro} & \textbf{sk} & \textbf{sv} & \textbf{ar} & \textbf{ca} & \textbf{hi} & \textbf{ru} & \textbf{uk} & \textbf{zh} \\
\midrule
\multicolumn{19}{c}{\textbf{\textit{Fully-open}}} \\
\midrule
\multicolumn{19}{l}{\textbf{\textit{European}}} \\
\addlinespace[2pt]
EuroLLM-9B-Base & 52.1 & 54.7 & 56.1 & 54.6 & 46.8 & 47.2 & 56.0 & 52.8 & 56.1 & 51.3 & 47.9 & 52.6 & 46.2 & 49.1 & 36.3 & 52.2 & 47.7 & 53.0 \\
EuroLLM-22B-Base & 54.7 & 57.0 & 57.2 & 57.4 & 49.7 & \textbf{\underline{50.3}} & 56.6 & 56.7 & 57.9 & 54.6 & 51.8 & 54.7 & 48.5 & 49.6 & \textbf{\underline{39.3}} & 54.5 & 49.5 & \textbf{\underline{55.5}} \\
Apertus-8B-Base & 54.0 & 56.7 & 57.9 & 57.9 & 51.5 & 48.6 & 59.3 & 55.7 & 58.1 & 48.7 & 50.3 & 54.5 & 46.9 & 51.9 & 36.6 & 55.1 & 51.3 & 51.8 \\
Apertus-70B-Base & \textbf{\underline{56.4}} & \textbf{\underline{58.2}} & \textbf{\underline{60.9}} & \textbf{\underline{59.8}} & \textbf{\underline{54.0}} & 48.7 & \textbf{\underline{60.1}} & \textbf{\underline{57.7}} & \textbf{\underline{60.9}} & \textbf{\underline{55.2}} & \textbf{\underline{52.0}} & \textbf{\underline{56.5}} & \textbf{\underline{50.0}} & \textbf{\underline{52.0}} & 37.1 & \textbf{\underline{55.4}} & \textbf{\underline{51.9}} & 53.2 \\
\addlinespace[4pt]
\multicolumn{19}{l}{\textbf{\textit{Non-European}}} \\
\addlinespace[2pt]
OLMo-3-7B-Base & 32.2 & 36.8 & 43.4 & 42.8 & 27.1 & 28.0 & 38.0 & 32.8 & 40.7 & 33.9 & 29.7 & 35.4 & 28.6 & 35.1 & 24.7 & 35.1 & 29.8 & 38.0 \\
OLMo-3-32B-Base & 46.6 & 52.3 & 58.1 & 54.9 & 40.2 & 36.1 & 55.0 & 49.3 & 56.4 & 47.4 & 38.4 & 48.3 & 40.9 & 45.7 & 33.6 & 48.5 & 40.2 & 52.4 \\
\midrule
\multicolumn{19}{c}{\textbf{\textit{Open-weights}}} \\
\midrule
\multicolumn{19}{l}{\textbf{\textit{European}}} \\
\addlinespace[2pt]
Mistral-3.2-24B-Base & 58.8 & \textbf{64.0} & 65.2 & \textbf{63.6} & 55.8 & 49.2 & \textbf{65.7} & 60.5 & 67.0 & 60.9 & 53.5 & 60.2 & 51.3 & \textbf{59.3} & 33.6 & 60.8 & 55.3 & 59.0 \\
\addlinespace[4pt]
\multicolumn{19}{l}{\textbf{\textit{Non-European}}} \\
\addlinespace[2pt]
Llama-3.1-8B-Base & 42.1 & 47.6 & 50.7 & 47.3 & 41.9 & 39.8 & 50.3 & 43.7 & 48.0 & 43.9 & 38.3 & 46.4 & 37.8 & 44.7 & 34.5 & 45.5 & 41.2 & 44.8 \\
Llama-3.3-70B-Base & 56.1 & 63.6 & 62.1 & 61.7 & 54.0 & 54.1 & 62.5 & 60.3 & 62.9 & 59.5 & 52.9 & 61.9 & 51.9 & 59.2 & \textbf{43.8} & 60.2 & 54.8 & 57.8 \\
Gemma-3-12B-Base & 59.0 & 61.0 & 64.4 & 60.8 & 55.7 & 52.5 & 61.7 & 60.2 & 65.4 & 58.7 & 55.8 & 60.6 & 52.8 & 57.1 & \textbf{43.8} & 58.6 & 54.5 & 59.2 \\
Gemma-3-27B-Base & \textbf{61.8} & 63.9 & \textbf{67.3} & \textbf{63.6} & \textbf{60.5} & \textbf{55.5} & 65.3 & \textbf{63.6} & \textbf{68.6} & \textbf{61.8} & \textbf{57.4} & \textbf{63.1} & \textbf{54.9} & 57.4 & 43.3 & \textbf{62.5} & \textbf{57.9} & \textbf{63.5} \\
Qwen-3-14B-Base & 52.7 & 59.3 & 61.3 & 58.0 & 50.4 & 49.8 & 61.4 & 56.3 & 59.5 & 54.8 & 52.7 & 53.5 & 47.9 & 52.7 & 41.2 & 56.4 & 50.0 & 59.3 \\
Qwen-3-30B-A3B-Base & 56.0 & 58.7 & 59.6 & 58.1 & 51.6 & 51.8 & 62.3 & 53.7 & 60.1 & 53.6 & 49.8 & 55.1 & 48.5 & 53.7 & 41.5 & 56.4 & 52.2 & 57.2 \\
\bottomrule
\end{tabular}}
\caption{Per-language performance on multilingual ARC-C. \textbf{Bold} indicates the best score per benchmark within each section (\textit{Fully-open} or \textit{Open-weights}). \underline{Underlined} indicates the best \textit{Fully-open} European score.}
\label{tab:base_results_marcc}
\end{table}

\clearpage

\section{Regex versus LLM-as-a-Judge}
\label{apd:regex_vs_llm}

In this appendix, we analyze the correlation between regex-based extraction, LLM-as-a-judge evaluations, and human judgments, providing evidence for why we relied on LLM-based assessment.

\subsection{Setup}

\paragraph{Models and tasks.}
To balance annotation cost with statistical rigor, we selected three models: Llama-3.3-70B, Qwen-3-32B, and Gemma-3-27B. We evaluated them on four tasks: MMLU, MMLU-Pro, GSM8K, and GPQA $\blacklozenge$, covering both simple tasks (MMLU, GSM8K) and more complex tasks (MMLU-Pro, GPQA $\blacklozenge$), as well as different output formats: letters (MMLU, MMLU-Pro, GPQA $\blacklozenge$) and numerical answers (GSM8K).

\paragraph{Prompting and parsing.}
To ensure consistent outputs, prompts were slightly adapted following \citet{hernándezcano2025apertus}. Each prompt concluded with the phrase \verb|"Answer with 'the answer is X'"| to encourage standardized responses, allowing reliable regex parsing. Regex extraction used the same functions as in \citet{hernándezcano2025apertus}, while LLM-as-a-judge evaluations followed the procedure and models described in \autoref{sec:evaluation}.

\paragraph{Annotations.}
To analyze the correlation between human judgments, regex parsing, and LLM-as-a-judge evaluations, we randomly sampled 100 examples from each dataset. For each model, a human annotator reviewed the question, the model’s generated answer, and the ground truth, marking whether the generated answer matched the ground truth (1 for match, 0 otherwise), resulting in a total of 1,200 annotations. Pearson correlation coefficients were then computed for both regex-human and LLM-human pairs.

\subsection{Results}

\begin{table}[!htbp]
\centering
\begin{tabular}{llccccc}
\toprule
\multirow{2}{*}{\textbf{Task}} & \multirow{2}{*}{\textbf{Model}} & \multicolumn{2}{c}{\textbf{Correlation}} & \multicolumn{3}{c}{\textbf{Accuracy}} \\
\cmidrule(lr){3-4} \cmidrule(lr){5-7}
 & & Regex-Human & LLM-Human & Regex & LLM & Human \\
\midrule
\multirow{3}{*}{MMLU} 
 & Llama-3.3-70B & 80.55 & 99.44 & 84.00 & 88.67 & 89.00 \\
 & Qwen-3-32B              & 81.48 & 97.89 & 79.00 & 84.00 & 85.00 \\
 & Gemma-3-27B          & 100.00 & 100.00 & 80.00 & 80.00 & 80.00 \\
\midrule
\multirow{3}{*}{MMLU-Pro} 
 & Llama-3.3-70B & 80.97 & 99.76 & 56.00 & 66.33 & 66.00 \\
 & Qwen-3-32B              & 72.95 & 98.90 & 59.00 & 73.67 & 73.00 \\
 & Gemma-3-27B          & 100.00 & 98.72 & 59.00 & 61.00 & 59.00 \\
\midrule
\multirow{3}{*}{GSM8K} 
 & Llama-3.3-70B & 33.95 & 99.29 & 57.00 & 92.33 & 92.00 \\
 & Qwen-3-32B              & 39.99 & 100.00 & 59.00 & 90.00 & 90.00 \\
 & Gemma-3-27B          & 58.56 & 100.00 & 82.00 & 93.00 & 93.00 \\
\midrule
\multirow{3}{*}{GPQA $\blacklozenge$} 
 & Llama-3.3-70B & 85.10 & 98.67 & 42.00 & 49.33 & 50.00 \\
 & Qwen-3-32B              & 84.82 & 96.73 & 54.00 & 64.00 & 62.00 \\
 & Gemma-3-27B        & 100.00 & 99.78 & 46.00 & 46.33 & 46.00 \\
\bottomrule
\end{tabular}
\caption{Comparison of regex-based and LLM-based evaluations in terms of correlation with human judgments and accuracy.}
\label{tab:regex_vs_llm}
\end{table}

\paragraph{LLM-as-a-judge aligns more closely with human judgments than regex-based parsing.}
\autoref{tab:regex_vs_llm} shows that, on average, LLM-based evaluation correlates far better with human judgments than regex-based methods. This difference arises because each model often formats its answers differently, and regex functions cannot reliably capture all variations (e.g., bold, italic, boxed text). While one could perform an extensive study to design the optimal regex function for each model, this would require substantial and tedious work that can be avoided by using LLM judges, which consistently achieve correlations above 96\% with human judgments.

\paragraph{Regex-based evaluation can affect performance rankings.}
Low correlation between regex and human judgments can lead to misleading evaluation outcomes. For instance, \autoref{tab:regex_vs_llm} shows that on MMLU, Gemma-3-27B appears to outperform Qwen-3-32B under regex-based evaluation but underperforms according to both LLM-based evaluation and human judgments. This discrepancy is largely due to Gemma adhering more strictly to formatting conventions. We argue that formatting should be considered only as part of evaluation (e.g., IFEval) and should not unduly influence other types of tasks when formatting differences are minor, such as bolding or italicization.

\section{Assessment Prompts}
\label{sec:assessment_prompt}

\begin{table}[!htbp]
\begin{tabular}{p{0.2\linewidth} p{0.75\linewidth}}
\toprule
\textbf{Task} & \textbf{Assessment Prompt} \\
\midrule
Default & 
\begin{minipage}[t]{\linewidth}
\begin{verbatim}
You are an evaluator. Your task is to determine 
whether the GENERATED ANSWER is equivalent in 
meaning to the GROUND TRUTH answer, given the 
QUESTION.
Respond only with "Answer: True" if the GENERATED 
ANSWER and GROUND TRUTH convey the same meaning, 
and "Answer: False" otherwise. 
Do not provide explanations.

QUESTION: 
{input}

GENERATED ANSWER: 
{generated_output}

GROUND TRUTH: 
{ground_truth}
\end{verbatim}
\end{minipage} \\
\midrule
IFEval &
\begin{minipage}[t]{\linewidth}
\begin{verbatim}
You are an evaluator. Your task is to determine 
whether the GENERATED ANSWER fully complies 
with the given INSTRUCTION.
Respond only with "Answer: True" if the GENERATED
ANSWER strictly follows the INSTRUCTION, and 
"Answer: False" otherwise. 
Do not provide explanations.

INSTRUCTION:
{input}

GENERATED ANSWER:
{generated_output}
\end{verbatim}
\end{minipage} \\
\bottomrule
\end{tabular}
\caption{Assessment prompts used for evaluating non-translation tasks with LLM-as-a-judge.}
\label{tab:assessment_prompts}
\end{table}

We provide the assessment prompts used for evaluating non-translation tasks with LLM-as-a-judge (\autoref{tab:assessment_prompts}). Since IFEval does not have a proper ground truth, it is evaluated using a different prompt that asks the judge to determine whether the generated output complies with the instructions provided in the input.

\end{document}